\documentclass[11pt]{article}

\PassOptionsToPackage{dvipsnames,table}{xcolor}\usepackage[final]{acl}

\usepackage{times}
\usepackage{latexsym}

\usepackage[T1]{fontenc}

\usepackage[utf8]{inputenc}

\usepackage{microtype}

\usepackage{inconsolata}

\usepackage{graphicx}
\usepackage{amsmath}
\usepackage{amssymb}
\usepackage{algorithm}
\usepackage{algorithmic}
\usepackage{booktabs}

\usepackage{multirow}
\usepackage{tabularx}

\title{Dynamic Alignment Compensation for Hallucination Mitigation in Large Vision-Language Models}

\author{
 \textbf{Kairong Yu\textsuperscript{1,*}} \quad
 \textbf{Zixin Zhu\textsuperscript{1,*}} \quad
 \textbf{Le Yu\textsuperscript{2}} \quad
 \textbf{Hongwei Wang\textsuperscript{1,\textdagger}}
\\
 \textsuperscript{1}Zhejiang University,
 \textsuperscript{2}Southeast University
\\
 \small{
    \textsuperscript{*}Equal contribution. \quad
   \textsuperscript{\textdagger}Corresponding author. \quad}
   \\
   \small{\textbf{Correspondence:} \href{mailto:hongweiwang@intl.zju.edu.cn}{hongweiwang@intl.zju.edu.cn}
 }
}

\begin{document}
\maketitle
\begin{abstract}
Large Vision-Language Models (LVLMs) remain prone to hallucinations, producing responses that are irrelevant or inconsistent with the multimodal input. Existing mitigation methods mainly rely on external supervision, output calibration, or attention regulation, leaving the internal representation dynamics of autoregressive generation underexplored. We identify an inference-time failure mode in which cross-modal representations degrade across decoder layers and drift across generation steps, destabilizing token prediction and increasing hallucination risk. We propose \emph{Dynamic Alignment Compensation} (DAC), a training-free inference-time method that detects representation divergence and selectively applies lightweight residual compensation. DAC combines Layer-wise Semantic Compensation to mitigate inter-layer degradation with Sequential Semantic Correction to constrain temporal drift. Experiments on nine hallucination-focused and general-purpose multimodal benchmarks across multiple LVLM backbones show that DAC consistently reduces hallucinations while maintaining strong overall performance.
\end{abstract}

\section{Introduction}
Large Vision-Language Models (LVLMs) have become a central paradigm for multimodal understanding, coupling visual inputs with autoregressive language generation~\cite{koh2023generating, tu2023sight}. 
They have been widely applied to visual question answering~\cite{antol2015vqa, shah2019kvqa, jia2025vqa2}, referring and reasoning-based segmentation~\cite{lai2024lisa, lan2024text4seg, zhu2025segagent}, and video retrieval~\cite{wang2024text, ventura2024covr}. 
Despite their progress, LVLMs remain prone to hallucinations, producing responses that are irrelevant to or inconsistent with the given multimodal input~\cite{gunjal2024detecting, lyu2025realrag}. 
Such failures undermine model reliability and can introduce substantial risks in high-stakes scenarios such as medical diagnosis~\cite{lin2025has} and autonomous driving~\cite{ding2024holistic}.

\begin{figure}[t]
    \centering
    \includegraphics[width=1.0\linewidth]{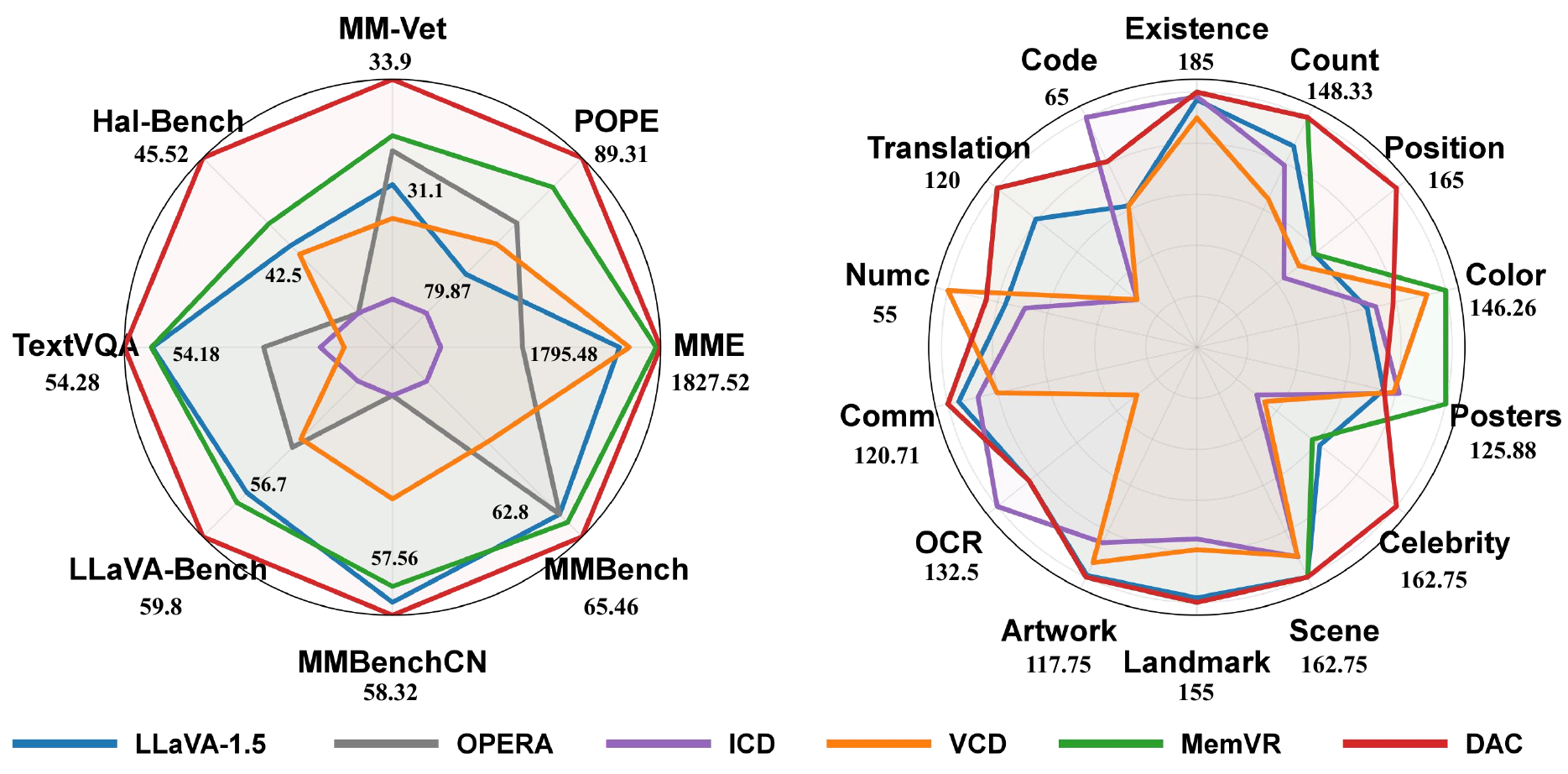}
    \caption{Performance comparison of different methods across multiple benchmarks. Hal-Bench denotes HallusionBench.}
    \label{fig:radar_chart}
\end{figure}

\begin{figure*}[t!]
    \centering
    \includegraphics[width=1.0\linewidth]{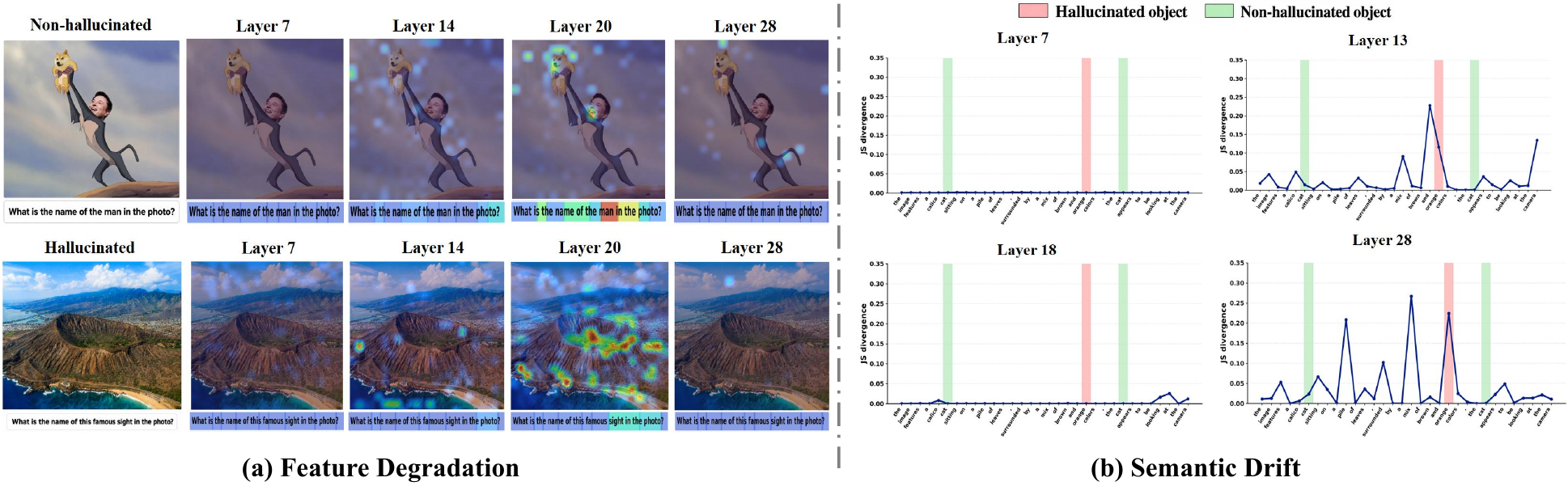}
    \caption{
Layer-wise and temporal JS-divergence patterns in hallucinated and non-hallucinated LVLM generation.
(a) Inter-layer JS divergence is projected onto image and text tokens at representative decoder layers. Hallucinated cases (bottom) show more frequent and stronger high-divergence regions than non-hallucinated cases (top), mainly in middle-to-late layers.
(b) Across decoding steps, hallucinated object tokens exhibit larger temporal drift than non-hallucinated object tokens, with a similar middle-to-late-layer concentration.
}
    \label{fig:motivation}
\end{figure*}

Existing hallucination mitigation methods can be broadly grouped into three paradigms.
Retrieval-augmented methods incorporate external evidence during generation~\cite{qu2025alleviating}, while fine-tuning-based methods rely on additional supervision or domain-specific data~\cite{yu2024rlhf}.
Both often require extra resources, such as retrieval databases, annotations, or model updates.
Training-free methods instead intervene during inference without updating model parameters.
Representative examples include logits-level contrastive decoding methods such as VCD and ICD~\cite{leng2024mitigating, wang2024mitigating}, attention-level methods such as OPERA~\cite{huang2024opera}, and representation-level methods such as MemVR~\cite{zou2024look}.

In this work, we analyze LVLM hallucination through hidden-state dynamics during autoregressive inference.
We identify two representation instabilities: inter-layer degradation across decoder layers and temporal drift across decoding steps.
To mitigate them, we propose \emph{Dynamic Alignment Compensation} (DAC), a training-free method that detects hidden-state divergence and applies lightweight residual compensation only when needed.
DAC integrates Layer-wise Semantic Compensation (LSC) for inter-layer degradation and Sequential Semantic Correction (SSC) for temporal drift.
Experiments on nine hallucination-focused and general-purpose multimodal benchmarks across multiple LVLM backbones show that DAC consistently reduces hallucinations while maintaining strong overall performance.

The main contributions of this work are:
\begin{itemize}
    \item We identify inter-layer degradation and temporal drift as two inference-time representation instabilities in LVLM generation.
    \item We propose DAC, a training-free decoding method for dynamic hidden-state compensation along layer and temporal dimensions.
    \item We evaluate DAC on nine multimodal benchmarks across multiple LVLM backbones, showing consistent hallucination reduction without training or external resources.
\end{itemize}

\section{Related Work}
\paragraph{Large Vision-Language Models.}
Large Vision-Language Models (LVLMs) extend large language models (LLMs) to multimodal inputs by coupling visual encoders with autoregressive language models through cross-modal interfaces.
Early systems such as BLIP-2~\cite{li2023blip} and LLaVA~\cite{liu2023visual} established this paradigm via query-based alignment and projection-based visual instruction tuning~\cite{yin2023lamm}.
Recent LVLMs further scale this design with stronger visual encoders, higher-resolution inputs, multi-image and video understanding, long-context modeling, and improved instruction alignment.
Representative models include the LLaVA family~\cite{liu2024llavanext, li2024llava}, the Qwen-VL family~\cite{wang2024qwen2, bai2025qwen3}, InternVL~\cite{chen2024internvl}, DeepSeek-VL2~\cite{wu2024deepseek}, and Kimi-VL~\cite{team2025kimi}.
In this work, we use LLaVA-1.5 and Qwen-VL as the main backbones, and further evaluate DAC on Qwen2.5-VL-3B, Qwen2.5-VL-7B, Qwen3-VL-4B, and LLaVA-NeXT-7B to examine its generality across model families, generations, and scales.

Despite this progress, LVLMs remain susceptible to hallucination, producing fluent and plausible responses that are inconsistent with the given multimodal context.
This problem is particularly prominent in open-ended generation, where outputs are not constrained by a fixed answer space and local errors may accumulate during autoregressive decoding.
Understanding the inference-time mechanisms behind such failures and developing reliable mitigation strategies therefore remain important challenges.

\paragraph{Hallucination in LVLMs.}
Hallucination in LVLMs has been studied from both diagnostic and mitigation perspectives~\cite{yin2024survey, zhou2023analyzing, bai2024hallucination}.
Prior studies link hallucinations to imperfect fine-grained cross-modal grounding, object co-occurrence bias, and over-reliance on language priors, which may cause generated responses to become inconsistent with the given multimodal context~\cite{rohrbach2018object, kim2023exposing}.
Supervision-based methods mitigate hallucinations through hallucination-aware instruction tuning~\cite{gunjal2024detecting}, post-hoc correction~\cite{zhou2023analyzing}, or reinforcement learning from human feedback~\cite{yu2024rlhf}.
Although effective, they typically require additional annotations, auxiliary models, or costly training.

Training-free methods instead intervene during inference and can be categorized by their intervention targets.
Attention-based methods, such as OPERA~\cite{huang2024opera}, EAH~\cite{zhang2024seeing}, and AGLA~\cite{an2025mitigating}, adjust attention distributions or visual-token usage during generation.
Logits- and decoding-based methods, such as VCD~\cite{leng2024mitigating}, ICD~\cite{wang2024mitigating}, SID~\cite{huo2024self}, and DeGF~\cite{zhang2025self}, calibrate token distributions through contrastive decoding, context-aware token selection, or generative feedback.
Representation-based methods, such as MemVR~\cite{zou2024look}, VISTA~\cite{pmlr-v267-li25ca}, and AIR~\cite{zhu2026look}, modify intermediate representations or salient internal features during generation.
Complementary to these inference-time interventions, we study hallucination from the perspective of hidden-state dynamics, focusing on representation divergence along decoder depth and successive decoding steps.

\begin{figure}[tb!]
    \centering
    \includegraphics[width=1.0\linewidth]{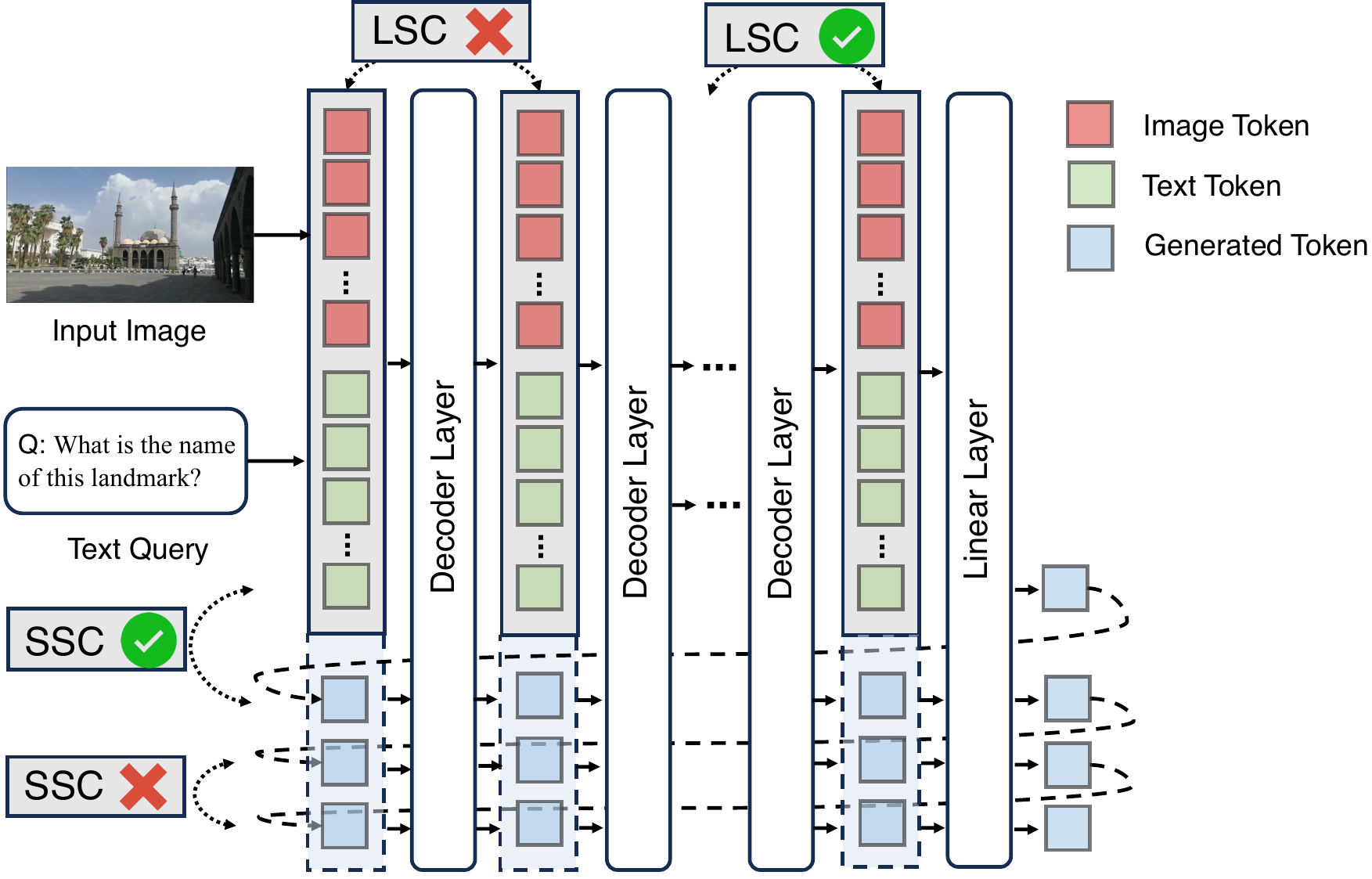}
    \caption{
Overview of Dynamic Alignment Compensation (DAC).
LSC compensates hidden states with large inter-layer representation divergence, while SSC constrains temporal drift of the current token using a layer-wise cache.
Green checks denote triggered updates; red crosses denote unchanged states.
}
    \label{fig:method}
\end{figure}

\section{Background and Motivation}
\label{sec:background_motivation}
\subsection{Autoregressive Inference in LVLMs}
\label{sec:background}

We consider an LVLM $\mathcal{M}_\Theta$ parameterized by $\Theta$, consisting of a visual encoder, a cross-modal projection module, and a Transformer-based autoregressive decoder.
Given an image $I$ and a textual query $Q$, the visual encoder and projection module convert $I$ into visual tokens $X_{\mathrm{vis}}$, while $Q$ is tokenized into text tokens $X_{\mathrm{txt}}$.
The initial multimodal sequence is
\begin{equation}
\label{eq:input_seq}
    X_0 = [X_{\mathrm{vis}}, X_{\mathrm{txt}}].
\end{equation}
At decoding step $t$, the decoder takes the extended sequence
$X_t = [X_0, y_{<t}]$, where $y_{<t}$ denotes previously generated tokens.
Let $H_{0,t}$ be the input embeddings of $X_t$.
The hidden states are updated layer by layer as
\begin{equation}
\label{eq:propagation}
    H_{l,t} = \mathcal{F}_l(H_{l-1,t}), \quad l = 1, \dots, L,
\end{equation}
where $\mathcal{F}_l(\cdot)$ denotes the $l$-th decoder layer.
The next-token distribution is computed from the final-layer hidden state at the current decoding position:
\begin{equation}
\label{eq:prediction}
    p_t = p_{\mathcal{M}_\Theta}(\cdot \mid I, Q, y_{<t})
    = \mathrm{softmax}\!\left(g_\Theta(h_{L,t})\right),
\end{equation}
where $h_{L,t}=H^{(\mathrm{cur})}_{L,t}$ and $g_\Theta(\cdot)$ is the output projection to the vocabulary space.
The next token is then sampled or selected from $p_t$.

Under this formulation, multimodal context is carried forward through the evolution of intermediate hidden states.
Therefore, representation instability during layer-wise propagation or across decoding steps can perturb the next-token distribution and increase the risk of hallucinated generation.

\subsection{Representation Divergence in Hallucinated Generation}
\label{sec:motivation}
Prior studies have linked LVLM hallucination to dataset bias, object co-occurrence, and language priors~\cite{zhou2023analyzing, leng2024mitigating}.
We provide a complementary inference-time perspective by examining how hidden representations evolve during autoregressive decoding.
Since next-token prediction depends on hidden states propagated through the decoder, abrupt representation changes along depth or decoding steps can perturb the decoding distribution.
We use token-level Jensen-Shannon (JS) divergence between feature-normalized hidden states as a lightweight measure of such representation shifts.

\paragraph{Inter-Layer Representation Degradation.}
Decoder layers should refine multimodal representations while preserving semantic continuity.
As shown in Fig.~\ref{fig:motivation}(a), non-hallucinated generation exhibits relatively sparse high-JS regions over image and text tokens, whereas hallucinated generation shows more frequent and stronger high-divergence regions.
These regions are not uniformly distributed across depth; they mainly emerge in middle-to-late decoder layers, while shallow and final layers show weaker separation.
This pattern suggests that hallucinated generation is associated with localized inter-layer degradation during hidden-state propagation.

\paragraph{Temporal Representation Drift.}
Hallucination is also reflected in the temporal evolution of hidden states.
Fig.~\ref{fig:motivation}(b) compares temporal JS divergence across decoding steps for hallucinated and non-hallucinated object tokens.
Hallucinated object tokens exhibit larger representation drift, again concentrated in middle-to-late decoder layers.
Together, these observations motivate DAC to compensate hidden-state instability along both layer and temporal dimensions.

\begin{table*}[tb!]
\centering
\small
\caption{Performance evaluation on MME, MM-Vet, MMBench, TextVQA, and MMBench-CN benchmarks.}
\label{tab:general-benchmarks}
\setlength{\tabcolsep}{3.8pt}
\renewcommand{\arraystretch}{1.05}
\begin{tabular}{c|lccccccc}
\hline
\multirow{2}{*}{Backbone} 
& \multirow{2}{*}{Method} 
& \multicolumn{3}{c}{MME} 
& MM-Vet 
& MMBench 
& TextVQA
& MMBench-CN \\ 
\cline{3-5} \cline{6-6} \cline{7-7} \cline{8-8} \cline{9-9}
& 
& Perception$\uparrow$
& Cognition$\uparrow$
& Overall$\uparrow$
& Score$\uparrow$
& Accuracy$\uparrow$
& Accuracy$\uparrow$
& Accuracy$\uparrow$ \\
\hline

\multirow{10}{*}{LLaVA-1.5}
& Regular 
& 1406.53 & 335.36 & 1741.89 & 31.10 & 62.80 & 54.18 & 57.56 \\

& OPERA   
& 1473.62 & 310.71 & 1784.34 & 32.00 & 62.80 & -- & -- \\

& ICD     
& 1341.63 & 295.36 & 1636.99 & 25.90 & 39.78 & 47.59 & 45.36 \\

& VCD     
& 1302.66 & 290.36 & 1593.02 & 30.20 & 54.21 & 45.71 & 51.46 \\

& MemVR   
& 1465.78 & 360.71 & 1826.49 & 32.40 & 63.75 & 54.22 & 56.62 \\

& VISTA   
& -- & -- & 1771.87 & -- & -- & -- & -- \\

& AGLA    
& 1472.66 & -- & -- & -- & -- & -- & -- \\

& DeGF    
& -- & -- & -- & -- & \textbf{65.50} & -- & -- \\

& AIR     
& -- & -- & -- & 32.00 & -- & -- & -- \\

& \cellcolor{gray!15}DAC
& \cellcolor{gray!15}\textbf{1501.22}
& \cellcolor{gray!15}\textbf{360.71}
& \cellcolor{gray!15}\textbf{1861.93}
& \cellcolor{gray!15}\textbf{33.90}
& \cellcolor{gray!15}65.46
& \cellcolor{gray!15}\textbf{54.38}
& \cellcolor{gray!15}\textbf{58.32} \\
\hline

\multirow{6}{*}{Qwen-VL-Chat}
& Regular 
& 1442.79 & 342.14 & 1784.93 & 49.00 & 56.53 & -- & -- \\

& OPERA   
& -- & -- & -- & -- & -- & -- & -- \\

& ICD     
& 1472.06 & \textbf{361.69} & 1833.75 & 29.37 & 13.32 & -- & -- \\

& VCD     
& 1403.17 & 317.89 & 1721.06 & 34.54 & 39.18 & -- & -- \\

& MemVR   
& 1473.45 & 347.50 & 1820.95 & 49.60 & 56.44 & -- & -- \\

& \cellcolor{gray!15}DAC
& \cellcolor{gray!15}\textbf{1509.16}
& \cellcolor{gray!15}354.64
& \cellcolor{gray!15}\textbf{1863.80}
& \cellcolor{gray!15}\textbf{50.00}
& \cellcolor{gray!15}\textbf{57.14}
& \cellcolor{gray!15}--
& \cellcolor{gray!15}-- \\
\hline
\end{tabular}
\end{table*}


\section{Methodology}
\label{sec:methodology}
Motivated by the representation-divergence patterns in Sec.~\ref{sec:motivation}, we propose \emph{Dynamic Alignment Compensation} (DAC), a training-free decoding method that stabilizes hidden-state propagation along both layer and temporal dimensions.
DAC consists of two components: Layer-wise Semantic Compensation (LSC), which compensates token states with large inter-layer divergence, and Sequential Semantic Correction (SSC), which constrains the step-wise drift of the current-token representation.

\subsection{Layer-wise Semantic Compensation}

At decoding step $t$, let $H_{l,t}\in\mathbb{R}^{n_t\times d}$ denote the hidden states output by decoder layer $l$ for the current sequence $X_t=[X_0,y_{<t}]$, where $n_t$ is the sequence length and $d$ is the hidden dimension.
For each token $i$ and layer $l\geq 2$, we normalize the adjacent-layer hidden states along the feature dimension:
\begin{equation}
    p_{l-1,t}^{(i)} = s\!\left(H_{l-1,t}^{(i)}\right),
    \quad
    p_{l,t}^{(i)} = s\!\left(H_{l,t}^{(i)}\right),
\end{equation}
where $s(\cdot)$ denotes feature-wise softmax.

For two normalized vectors $p$ and $q$, let $m=\tfrac{1}{2}(p+q)$.
We define
\begin{equation}
\mathrm{JS}(p\|q)
=
\tfrac{1}{2}\mathrm{KL}(p\|m)
+
\tfrac{1}{2}\mathrm{KL}(q\|m).
\label{eq:js}
\end{equation}

The token-level inter-layer divergence is then computed as
\begin{equation}
    D_{l,t}^{(i)}
    =
    \mathrm{JS}\!\left(
    p_{l-1,t}^{(i)}
    \,\|\, 
    p_{l,t}^{(i)}
    \right).
\end{equation}

LSC compensates only tokens whose divergence exceeds a threshold:
\begin{equation}
    \bar{H}_{l,t}^{(i)}
    =
    H_{l,t}^{(i)}
    +
    \alpha\,\mathbf{1}\!\left[D_{l,t}^{(i)}>\gamma\right]
    H_{l-1,t}^{(i)} ,
\end{equation}
where $\gamma$ is the inter-layer divergence threshold, $\alpha$ controls the compensation strength, and $\mathbf{1}[\cdot]$ is the indicator function.
Tokens below the threshold remain unchanged.
The compensated state $\bar{H}_{l,t}$ is used as the layer output.

\subsection{Sequential Semantic Correction}

While LSC stabilizes hidden-state propagation across adjacent layers, autoregressive decoding may still introduce drift across successive generation steps.
SSC constrains this drift using a lightweight layer-wise temporal cache.
After LSC, let $\bar{h}_{l,t}=\bar{H}_{l,t}^{(\mathrm{cur})}$ denote the current-token hidden state at layer $l$.
The cache $c_{l,t-1}$ stores the corrected current-token state from the previous decoding step.
When the cache is available, we compute
\begin{equation}
    q_{l,t} = s(\bar{h}_{l,t}),
    \quad
    q_{l,t-1} = s(c_{l,t-1}),
\end{equation}
and measure temporal drift by
\begin{equation}
    \Delta_{l,t}
    =
    \mathrm{JS}\!\left(
    q_{l,t}
    \,\|\, 
    q_{l,t-1}
    \right).
\end{equation}

If the drift exceeds a threshold, SSC moves the current state toward the cached state:
\begin{equation}
    \tilde{h}_{l,t}
    =
    \bar{h}_{l,t}
    +
    \beta\,\mathbf{1}\!\left[\Delta_{l,t}>\tau\right]
    \left(c_{l,t-1}-\bar{h}_{l,t}\right),
\end{equation}
where $\tau$ is the temporal drift threshold and $\beta\in[0,1]$ controls the correction strength.
The corrected state $\tilde{h}_{l,t}$ is written back to the current-token position and stored as $c_{l,t}$ for the next decoding step.
When no previous cache is available, SSC is skipped and the current state is directly cached.

At each decoding step, DAC applies LSC before SSC, so temporal correction operates on layer-stabilized hidden states.
The complete inference pipeline is illustrated in Fig.~\ref{fig:method} and summarized in Algorithm~\ref{alg:dac}.

\begin{table*}[t!]
\centering
\small
\caption{Performance evaluation on the POPE benchmark.}
\label{tab:pope}
\setlength{\tabcolsep}{4.8pt}
\renewcommand{\arraystretch}{1.06}
\begin{tabular}{lcccccccc}
\hline
Method 
& \multicolumn{2}{c}{Random} 
& \multicolumn{2}{c}{Popular} 
& \multicolumn{2}{c}{Adversarial} 
& \multicolumn{2}{c}{Average} \\ 
\cline{2-9}
& Accuracy$\uparrow$ & F1-score$\uparrow$
& Accuracy$\uparrow$ & F1-score$\uparrow$
& Accuracy$\uparrow$ & F1-score$\uparrow$
& Accuracy$\uparrow$ & F1-score$\uparrow$ \\
\hline

LLaVA-1.5 
& 83.49 & 82.28 
& 79.98 & 79.34 
& 76.03 & 76.26 
& 79.83 & 79.29 \\

+OPERA 
& 87.53 & 86.45 
& 84.21 & 83.50 
& 80.88 & 80.69 
& 84.21 & 83.55 \\

+ICD 
& 84.87 & 83.27 
& 82.93 & 81.45 
& 81.07 & 79.96 
& 82.96 & 81.56 \\

+VCD 
& 86.84 & 86.83 
& 82.65 & 83.37 
& 77.31 & 79.28 
& 82.27 & 83.16 \\

+MemVR 
& 88.50 & 87.34 
& 87.10 & 86.01 
& \textbf{85.20} & 84.28 
& 86.93 & 85.88 \\

+AGLA 
& 88.54 & 87.71 
& 85.14 & 84.68 
& 81.13 & 81.36 
& 84.94 & 84.58 \\

+DeGF 
& 89.03 & 88.74 
& 86.63 & 86.28 
& 81.63 & 80.59 
& 85.76 & 85.65 \\

+VISTA 
& -- & -- 
& -- & -- 
& -- & -- 
& 86.15 & 86.29 \\

\rowcolor{gray!15}
+DAC 
& \textbf{89.31} & \textbf{88.88} 
& \textbf{87.73} & \textbf{87.10} 
& 84.93 & \textbf{84.91} 
& \textbf{87.32} & \textbf{86.96} \\

\hline
\end{tabular}
\end{table*}

\section{Experiments}
\subsection{Experiment Setup}
We evaluate DAC on nine multimodal benchmarks covering both hallucination-focused and general-purpose evaluation.
The hallucination-focused suite includes POPE~\cite{li2023evaluating}, CHAIR~\cite{rohrbach2018object}, and HallusionBench~\cite{guan2024hallusionbench}.
The general-purpose suite includes MME~\cite{fu2025mme}, MM-Vet~\cite{yu2023mm}, MMBench~\cite{liu2024mmbench}, MMBench-CN~\cite{liu2024mmbench}, TextVQA~\cite{singh2019towards}, and LLaVA-Bench~\cite{liu2023visual}.
Unless otherwise specified, all evaluations follow the official protocols.
Detailed settings, metrics, and additional results are provided in Appendix~\ref{sec:benchmarks}.

\subsection{Results on General-Purpose Benchmarks}
Tables~\ref{tab:general-benchmarks} and~\ref{tab:llavabench} report results on general-purpose multimodal benchmarks.
DAC consistently improves the vanilla decoding baseline on both LLaVA-1.5 and Qwen-VL-Chat across the reported evaluations.
On MME, DAC increases the overall score by 120.04 points on LLaVA-1.5 and 78.87 points on Qwen-VL-Chat, with gains in both perception and cognition.
It also improves MM-Vet and MMBench on both backbones, achieving 33.90/65.46 on LLaVA-1.5 and 50.00/57.14 on Qwen-VL-Chat, respectively.
On LLaVA-1.5, DAC further improves TextVQA and MMBench-CN, indicating that hidden-state compensation does not compromise OCR-oriented or cross-lingual evaluation.

Table~\ref{tab:llavabench} further evaluates open-ended visual instruction following.
DAC improves the average LLaVA-Bench score from 57.03 to 61.55 on LLaVA-1.5 and from 71.53 to 76.08 on Qwen-VL-Chat.
Overall, DAC maintains strong general-purpose multimodal performance while remaining competitive with representative logits-, attention-, and representation-level mitigation methods.

\begin{table*}[t]
\centering
\small
\caption{Generalization to recent LVLM backbones on general-purpose multimodal benchmarks.}
\label{tab:additional-backbones}
\setlength{\tabcolsep}{3.6pt}
\renewcommand{\arraystretch}{1.08}
\begin{tabular}{@{}llccccc@{}}
\hline
Backbone & Method
& \shortstack{MM-Vet\\(Score)$\uparrow$}
& \shortstack{MMBench\\(Acc.)$\uparrow$}
& \shortstack{TextVQA\\(Acc.)$\uparrow$}
& \shortstack{MMBench-CN\\(Acc.)$\uparrow$}
& \shortstack{LLaVA-Bench\\(Score)$\uparrow$} \\
\hline

\multirow{5}{*}{Qwen2.5-VL-3B}
& Regular & 62.1 & 78.35 & 77.41 & 77.06 & 82.37 \\
& +MemVR  & 60.2 & 77.83 & 77.45 & 75.51 & 83.42 \\
& +ICD    & 20.9 & 75.94 & 77.09 & 73.71 & 78.29 \\
& +VCD    & 57.4 & 75.17 & 72.76 & 74.57 & 80.38 \\
& \cellcolor{gray!15}+DAC
& \cellcolor{gray!15}\textbf{63.6}
& \cellcolor{gray!15}\textbf{78.92}
& \cellcolor{gray!15}\textbf{77.82}
& \cellcolor{gray!15}\textbf{77.41}
& \cellcolor{gray!15}\textbf{86.60} \\
\hline

\multirow{5}{*}{Qwen2.5-VL-7B}
& Regular & 67.5 & 83.50 & 77.86 & 81.10 & 89.85 \\
& +MemVR  & 68.6 & 83.33 & 75.65 & 80.49 & 89.22 \\
& +ICD    & 19.4 & 81.87 & 75.65 & 79.37 & 81.73 \\
& +VCD    & 65.7 & 81.52 & 76.02 & 79.55 & 85.04 \\
& \cellcolor{gray!15}+DAC
& \cellcolor{gray!15}\textbf{69.3}
& \cellcolor{gray!15}\textbf{83.91}
& \cellcolor{gray!15}\textbf{77.94}
& \cellcolor{gray!15}\textbf{81.37}
& \cellcolor{gray!15}\textbf{95.18} \\
\hline

\multirow{5}{*}{Qwen3-VL-4B}
& Regular & 68.0 & 84.10 & 78.13 & 81.27 & 95.62 \\
& +MemVR  & 66.6 & 84.10 & 77.84 & 81.18 & 94.10 \\
& +ICD    & 6.1  & 75.94 & 78.37 & 80.49 & 91.48 \\
& +VCD    & 8.6  & 83.91 & 77.05 & 80.75 & 88.03 \\
& \cellcolor{gray!15}+DAC
& \cellcolor{gray!15}\textbf{70.6}
& \cellcolor{gray!15}\textbf{84.68}
& \cellcolor{gray!15}\textbf{78.49}
& \cellcolor{gray!15}\textbf{81.36}
& \cellcolor{gray!15}\textbf{112.20} \\
\hline

\multirow{5}{*}{LLaVA-NeXT-7B}
& Regular & 41.4 & 67.86 & 61.33 & 60.56 & 54.28 \\
& +MemVR  & 40.3 & 67.96 & \textbf{61.41} & 60.56 & 57.07 \\
& +ICD    & 19.2 & 62.80 & 60.28 & 55.49 & 49.82 \\
& +VCD    & 25.8 & 63.40 & 54.64 & 56.26 & 52.56 \\
& \cellcolor{gray!15}+DAC
& \cellcolor{gray!15}\textbf{41.9}
& \cellcolor{gray!15}\textbf{68.44}
& \cellcolor{gray!15}\textbf{61.41}
& \cellcolor{gray!15}\textbf{60.85}
& \cellcolor{gray!15}\textbf{58.88} \\
\hline
\end{tabular}
\end{table*}

\begin{table}[t]
\centering
\small
\caption{Performance evaluation on CHAIR. $C_s$ and $C_i$ denote sentence-level and instance-level hallucination rates, respectively; Len. denotes average response length.}
\label{tab:chair}
\setlength{\tabcolsep}{4.2pt}
\renewcommand{\arraystretch}{1.05}
\begin{tabular}{lcccc}
\hline
Method 
& $C_s\downarrow$ 
& $C_i\downarrow$ 
& Recall$\uparrow$ 
& Len. \\
\hline

LLaVA-1.5 
& 47.6 & 13.3 & 80.6 & 99.7 \\
+OPERA 
& 47.6 & 13.5 & 79.0 & 93.2 \\
+ICD 
& 56.2 & 16.3 & 16.3 & 103.4 \\
+VCD 
& 55.0 & 15.8 & 77.4 & 101.2 \\
+MemVR 
& 46.6 & 13.0 & \textbf{80.8} & 99.6 \\
\rowcolor{gray!15}
+DAC 
& \textbf{28.2} & \textbf{7.3} & 72.0 & 88.9 \\
\hline

Qwen-VL-Chat 
& 6.8 & 5.3 & \textbf{53.4} & 17.6 \\
+OPERA 
& -- & -- & -- & -- \\
+ICD 
& 18.4 & 14.3 & 37.6 & 48.1 \\
+VCD 
& 13.0 & 12.3 & 47.9 & 115.7 \\
+MemVR 
& 4.8 & 3.3 & 52.3 & 15.0 \\
\rowcolor{gray!15}
+DAC 
& \textbf{3.8} & \textbf{2.5} & 50.9 & 14.3 \\
\hline
\end{tabular}
\end{table}

\begin{table}[tb]
\centering
\small
\caption{Performance evaluation on LLaVA-Bench. C., D., Cp., and Avg. denote Conversation, Detail, Complex, and Average, respectively.}
\label{tab:llavabench}
\setlength{\tabcolsep}{2.5pt}
\renewcommand{\arraystretch}{1.06}
\begin{tabular}{@{}c@{\hspace{4pt}}l@{\hspace{5pt}}ccccc@{}}
\hline
Backbone & Method 
& C.$\uparrow$ 
& D.$\uparrow$ 
& Cp.$\uparrow$ 
& All$\uparrow$  
& Avg.$\uparrow$ \\ 
\hline

\multirow{5}{*}{LLaVA-1.5} 
& Regular & 51.2 & 50.3 & 70.0 & 56.6 & 57.03 \\
& ICD     & 49.3 & 46.9 & 63.1 & 54.9 & 53.55 \\
& VCD     & 50.0 & 47.2 & 62.0 & 54.7 & 53.48 \\
& MemVR   & 53.4 & 50.7 & 71.1 & 57.4 & 60.53 \\
& \cellcolor{gray!15}DAC     
& \cellcolor{gray!15}\textbf{57.5} 
& \cellcolor{gray!15}\textbf{55.8}   
& \cellcolor{gray!15}\textbf{73.1}    
& \cellcolor{gray!15}\textbf{59.8} 
& \cellcolor{gray!15}\textbf{61.55} \\ 
\hline

\multirow{5}{*}{Qwen-VL-Chat}   
& Regular & 68.9 & 61.6 & \textbf{82.3} & 73.3 & 71.53 \\
& ICD     & 71.7 & 61.9 & 74.7 & 70.6 & 69.73 \\
& VCD     & 33.8 & 23.9 & 30.2 & 29.6 & 29.38 \\
& MemVR   & 69.4 & 62.8 & \textbf{82.3} & 73.7 & 72.05 \\
& \cellcolor{gray!15}DAC     
& \cellcolor{gray!15}\textbf{77.3} 
& \cellcolor{gray!15}\textbf{69.6}   
& \cellcolor{gray!15}81.5    
& \cellcolor{gray!15}\textbf{75.9} 
& \cellcolor{gray!15}\textbf{76.08} \\ 
\hline
\end{tabular}
\end{table}

\subsection{Results on Hallucination Benchmarks}
\label{sec:hallucination_results}

Tables~\ref{tab:pope}, \ref{tab:chair}, and~\ref{tab:hallusionbench} report results on POPE, CHAIR, and HallusionBench.
On POPE, DAC achieves the best average accuracy and F1, improving LLaVA-1.5 from 79.83 to 87.32 and from 79.29 to 86.96, respectively.
It obtains the best results on the Random and Popular splits, and the highest F1 on the Adversarial split, showing consistent reduction of object-level hallucination.

On CHAIR, DAC reduces $C_{\mathrm{s}}$/$C_{\mathrm{i}}$ from 47.6/13.3 to 28.2/7.3 on LLaVA-1.5, and from 6.8/5.3 to 3.8/2.5 on Qwen-VL-Chat.
Although recall decreases, the length-normalized analysis in Appendix~\ref{sec:app_chair_norm} shows that DAC also lowers hallucinated object mentions per generated token, indicating that the improvement is not merely due to shorter outputs.
On HallusionBench, DAC improves qACC from 8.13 to 17.05 and aACC from 41.5 to 45.5, while achieving the best easyA score.
Overall, DAC reduces hallucinations across complementary evaluation protocols, suggesting that dynamic hidden-state compensation improves generation reliability beyond a single benchmark setting.

\begin{table*}[t]
\centering
\small
\caption{Comparison on MME and POPE across recent LVLM backbones.MME reports perception, cognition, and overall scores; POPE reports average accuracy and F1 over three splits.
Best or tied-best results within each backbone block are bolded.}
\label{tab:additional-backbones-mme-pope}
\setlength{\tabcolsep}{5.0pt}
\renewcommand{\arraystretch}{1.05}
\begin{tabular}{@{}llccccc@{}}
\hline
Backbone & Method
& MME(Perception)$\uparrow$
& MME(Cognition)$\uparrow$
& MME(Overall)$\uparrow$
& POPE(Acc.)$\uparrow$
& POPE(F1)$\uparrow$ \\
\hline

\multirow{5}{*}{Qwen2.5-VL-3B}
& Regular & 1518.15 & 621.78 & 2139.93 & 87.38 & 86.20 \\
& +MemVR  & 1499.00 & \textbf{625.35} & 2124.35 & 87.51 & 86.36 \\
& +ICD    & 1473.30 & 571.42 & 2044.72 & 87.31 & 86.08 \\
& +VCD    & 1512.03 & 595.35 & 2107.38 & 86.91 & 85.80 \\
& \cellcolor{gray!15}+DAC
& \cellcolor{gray!15}\textbf{1528.00}
& \cellcolor{gray!15}621.78
& \cellcolor{gray!15}\textbf{2149.78}
& \cellcolor{gray!15}\textbf{87.67}
& \cellcolor{gray!15}\textbf{86.43} \\
\hline

\multirow{5}{*}{Qwen2.5-VL-7B}
& Regular & 1692.78 & 624.28 & 2317.06 & 87.67 & 86.54 \\
& +MemVR  & \textbf{1702.00} & 622.14 & 2324.14 & 87.68 & 86.56 \\
& +ICD    & 1666.24 & 612.50 & 2278.74 & 87.73 & 86.68 \\
& +VCD    & 1596.66 & 611.07 & 2207.73 & \textbf{88.05} & \textbf{87.13} \\
& \cellcolor{gray!15}+DAC
& \cellcolor{gray!15}1699.00
& \cellcolor{gray!15}\textbf{629.65}
& \cellcolor{gray!15}\textbf{2328.65}
& \cellcolor{gray!15}87.80
& \cellcolor{gray!15}86.73 \\
\hline

\multirow{5}{*}{Qwen3-VL-4B}
& Regular & 1711.75 & \textbf{609.64} & \textbf{2321.39} & 89.53 & 88.95 \\
& +MemVR  & \textbf{1720.99} & 581.42 & 2302.41 & 89.28 & 86.43 \\
& +ICD    & 1699.48 & 596.42 & 2295.90 & 88.96 & 88.33 \\
& +VCD    & 1682.90 & 606.78 & 2289.68 & 89.37 & 88.91 \\
& \cellcolor{gray!15}+DAC
& \cellcolor{gray!15}1711.75
& \cellcolor{gray!15}608.21
& \cellcolor{gray!15}2319.96
& \cellcolor{gray!15}\textbf{89.66}
& \cellcolor{gray!15}\textbf{89.03} \\
\hline

\multirow{5}{*}{LLaVA-NeXT-7B}
& Regular & 1519.29 & 322.50 & 1841.79 & 87.50 & 86.42 \\
& +MemVR  & 1501.55 & 308.57 & 1810.12 & \textbf{87.71} & \textbf{86.70} \\
& +ICD    & 1413.27 & 331.07 & 1744.34 & 86.25 & 85.09 \\
& +VCD    & 1369.61 & \textbf{340.71} & 1710.32 & 86.06 & 85.06 \\
& \cellcolor{gray!15}+DAC
& \cellcolor{gray!15}\textbf{1521.08}
& \cellcolor{gray!15}332.14
& \cellcolor{gray!15}\textbf{1853.22}
& \cellcolor{gray!15}87.45
& \cellcolor{gray!15}86.37 \\
\hline
\end{tabular}
\end{table*}

\subsection{Generalization Across Recent LVLM Backbones and Scales}
\label{sec:recent_backbones}

We further evaluate DAC on Qwen2.5-VL-3B, Qwen2.5-VL-7B, Qwen3-VL-4B, and LLaVA-NeXT-7B to examine its transferability across recent LVLM families and model scales.
As shown in Table~\ref{tab:additional-backbones}, DAC improves the vanilla baseline on every reported general-purpose metric across the four backbones.
On MM-Vet, DAC brings gains of 1.5, 1.8, 2.6, and 0.5 points, respectively.
The improvements are more pronounced on LLaVA-Bench, where DAC increases the scores by 4.23, 5.33, 16.58, and 4.60 points.
Compared with MemVR, ICD, and VCD, DAC achieves the best or tied-best result in every entry of Table~\ref{tab:additional-backbones}, indicating strong generalization beyond the main backbones.

Table~\ref{tab:additional-backbones-mme-pope} further evaluates DAC on MME and POPE.
DAC improves MME Overall by 9.85, 11.59, and 11.43 points on Qwen2.5-VL-3B, Qwen2.5-VL-7B, and LLaVA-NeXT-7B, respectively, while showing only a slight decrease on Qwen3-VL-4B.
On POPE, DAC improves average accuracy/F1 by 0.29/0.23, 0.13/0.19, and 0.13/0.08 on the two Qwen2.5-VL variants and Qwen3-VL-4B, with nearly unchanged results on LLaVA-NeXT-7B.
Overall, these results show that DAC remains effective across newer architectures and different parameter scales, supporting dynamic hidden-state compensation as a broadly applicable inference-time mechanism.

\begin{table}[tb]
\centering
\small
\caption{Performance evaluation on HallusionBench.}
\label{tab:hallusionbench}
\setlength{\tabcolsep}{3.2pt}
\renewcommand{\arraystretch}{1.06}
\begin{tabular}{lccccc}
\hline
Method 
& fACC$\uparrow$ 
& qACC$\uparrow$  
& easyA$\uparrow$ 
& hardA$\uparrow$ 
& aACC$\uparrow$ \\ 
\hline

LLaVA-1.5 
& \textbf{17.9} & 8.13 & 36.0 & 36.7 & 41.5 \\

+OPERA    
& 16.2 & 5.49 & 37.6 & 35.4 & 41.2 \\

+ICD      
& 13.9 & 8.35 & 36.9 & 33.5 & 38.2 \\

+VCD      
& 13.9 & 11.4 & 33.0 & 34.7 & 41.1 \\

+MemVR    
& \textbf{17.9} & 9.01 & 36.9 & \textbf{37.7} & 42.5 \\

\rowcolor{gray!15}
+DAC      
& 14.2 & \textbf{17.05} & \textbf{44.4} & 37.3 & \textbf{45.5} \\

\hline
\end{tabular}
\end{table}

\subsection{Ablation Studies}
\label{sec:ablation}
\paragraph{Effect of LSC and SSC.}
Table~\ref{tab:ablation-lsc-ssc} ablates the two components of DAC on MM-Vet with LLaVA-1.5.
Both components improve the vanilla baseline, raising the total score from 31.1 to 33.0 with SSC and to 33.3 with LSC.
SSC mainly benefits OCR and spatial reasoning, while LSC yields broader gains across recognition, OCR, knowledge, generation, and spatial reasoning.
Combining LSC and SSC achieves the best total score of 33.9, with the strongest OCR, knowledge, and spatial scores.
These results suggest that layer-wise compensation and temporal correction provide complementary benefits for stabilizing hidden-state dynamics during inference.

\begin{table}[tb!]
\centering
\small
\caption{Ablation results of LSC and SSC.}
\label{tab:ablation-lsc-ssc}
\setlength{\tabcolsep}{1.35pt}
\renewcommand{\arraystretch}{1.06}
\begin{tabular}{@{}cc@{\hspace{3pt}}ccccccc@{}}
\hline
LSC & SSC & Rec. & OCR & Kno. & Gen. & Spa. & Math & Total \\
\hline
$\times$     & $\times$     & 36.4 & 20.1 & 22.3 & 22.2 & 24.8 & 7.7 & 31.1 \\
$\times$     & $\checkmark$ & 37.2 & 23.5 & 23.3 & 22.6 & 28.9 & 7.7 & 33.0 \\
$\checkmark$ & $\times$     & 37.9 & 23.6 & 24.6 & \textbf{24.5} & 28.3 & 7.7 & 33.3 \\
$\checkmark$ & $\checkmark$ & \textbf{37.9} & \textbf{25.1} & \textbf{24.9} & 24.1 & \textbf{30.1} & \textbf{7.7} & \textbf{33.9} \\
\hline
\end{tabular}
\end{table}

\begin{figure*}[tb!]
    \centering
    \includegraphics[width=1.0\linewidth]{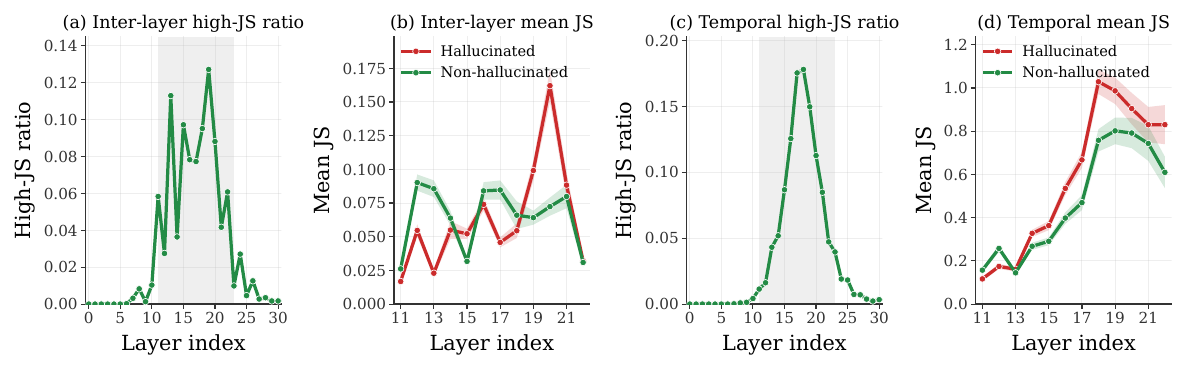}
    \caption{
JS-divergence diagnostics on CHAIR with LLaVA-1.5.
(a),(b) show the inter-layer high-JS ratio and mean JS across decoder layers, while (c),(d) show the corresponding temporal statistics across decoding steps.
Mean-JS curves compare hallucinated and non-hallucinated object tokens.
Gray regions highlight the middle-to-late layers where high-JS outliers concentrate, and shaded bands denote standard error.
}
\label{fig:js_diagnostic}
\end{figure*}

\paragraph{Hallucination-related JS divergence.}
The case visualization in Fig.~\ref{fig:motivation} provides an intuitive example of the representation instability associated with hallucinated generation.
In the layer-wise view, hallucinated generation exhibits denser and stronger high-JS regions over both image and text tokens than non-hallucinated generation, with the most visible changes appearing in middle-to-late decoder layers.
The temporal view shows a consistent trend: tokens corresponding to hallucinated objects undergo larger step-wise drift than tokens corresponding to non-hallucinated objects.
These observations motivate a dataset-level analysis to examine whether such patterns persist beyond individual examples.

We conduct this analysis on CHAIR with LLaVA-1.5.
For inter-layer divergence, Fig.~\ref{fig:js_diagnostic}(a) reports the high-JS ratio, which measures how frequently tokens exhibit large changes between adjacent decoder layers.
The outliers are concentrated in middle-to-late layers rather than being uniformly distributed across depth.
Fig.~\ref{fig:js_diagnostic}(b) further compares the mean JS of hallucinated and non-hallucinated object tokens within this region.
Hallucinated object tokens show a clear late-layer peak, indicating not only more frequent instability but also stronger representation shifts.

The temporal statistics show the same tendency.
Fig.~\ref{fig:js_diagnostic}(c) measures high-JS ratios between consecutive decoding steps and again identifies middle-to-late layers as the main unstable region.
Fig.~\ref{fig:js_diagnostic}(d) shows that hallucinated object tokens have consistently larger temporal JS divergence than non-hallucinated ones, suggesting that hallucinated generation is associated with accumulated drift along the decoding trajectory.
Together, the case-level visualization and dataset-level diagnostics support the design of DAC: LSC targets localized inter-layer instability, while SSC constrains temporal drift during autoregressive decoding.

\begin{table}[t]
\centering
\small
\caption{
Inference efficiency on MM-Vet with LLaVA-1.5.
Best or tied-best values are bolded.
}
\label{tab:inference-efficiency}
\setlength{\tabcolsep}{2.4pt}
\renewcommand{\arraystretch}{1.08}
\begin{tabular}{@{}lcccc@{}}
\hline
Method
& \shortstack{Latency\\(ms/token)$\downarrow$}
& \shortstack{Throughput\\(tokens/ms)$\uparrow$}
& \shortstack{T/80 Tokens\\(ms)$\downarrow$}
& \shortstack{Memory\\(MB)$\downarrow$} \\
\hline
Greedy
& \textbf{36.8} & \textbf{0.027} & \textbf{2946.91} & 13797 \\
Sampling
& 37.1 & \textbf{0.027} & 2972.99 & 13797 \\
MemVR
& \textbf{36.8} & \textbf{0.027} & 2949.12 & 13800 \\
ICD
& 79.0 & 0.013 & 6320.52 & \textbf{13778} \\
OPERA
& 137.1 & 0.008 & 11501.57 & 21110 \\
VCD
& 76.1 & 0.013 & 6085.18 & 13781 \\
\rowcolor{gray!15}
DAC
& 37.1 & 0.026 & 2975.45 & 13797 \\
\hline
\end{tabular}
\end{table}

\paragraph{Inference efficiency.}
We further evaluate the inference cost of DAC on MM-Vet with LLaVA-1.5.
As shown in Table~\ref{tab:inference-efficiency}, DAC introduces negligible overhead over standard decoding.
Compared with greedy decoding, DAC increases per-token latency only from 36.8 to 37.1 ms/token, while keeping the same peak memory usage of 13,797 MB.
Its throughput remains close to standard decoding, decreasing slightly from 0.027 to 0.026 tokens/ms.
For generating 80 tokens, DAC takes 2975.45 ms, which is comparable to greedy decoding and sampling, and substantially faster than ICD, VCD, and OPERA.
Specifically, DAC reduces the 80-token generation time by 52.9\%, 51.1\%, and 74.1\% compared with ICD, VCD, and OPERA, respectively.
These results show that DAC improves generation reliability with little additional inference cost, without relying on extra decoding branches or auxiliary models.

\section{Conclusion}
We introduced \emph{Dynamic Alignment Compensation} (DAC), a training-free inference-time method for mitigating LVLM hallucinations by stabilizing hidden-state dynamics.
DAC detects inter-layer representation degradation and temporal representation drift, and applies lightweight compensation through Layer-wise Semantic Compensation and Sequential Semantic Correction.
Experiments on nine hallucination-focused and general-purpose multimodal benchmarks show that DAC consistently reduces hallucinations while maintaining strong overall performance across multiple LVLM backbones and model scales.
Further analysis confirms the complementarity of its two components and shows that DAC incurs negligible inference overhead.
These results highlight dynamic hidden-state compensation as a practical direction for improving the reliability of LVLM generation.

\section{Limitations}
This work focuses on training-free hallucination mitigation for image-based LVLMs. Although DAC is evaluated on multiple hallucination-focused and general-purpose benchmarks across several open-source backbones, its effectiveness on closed-source models, video or multi-image inputs, long-context multimodal generation, and highly specialized domains remains to be further studied.
DAC uses feature-wise softmax and Jensen-Shannon divergence as lightweight signals for detecting representation instability. While these signals are empirically effective, other distance measures or adaptive criteria may further improve robustness. In addition, we use fixed thresholds and compensation coefficients across all datasets and models to avoid dataset-specific tuning; future work may explore input-adaptive parameter selection.
Finally, DAC reduces hallucination risk but does not guarantee factual correctness. For high-stakes applications, additional verification and safety assessment are still required.

\section*{Acknowledgments}
This work was supported by the National Key Research and Development Program of China (2024YFF0907802 and 2024YFF0907803), the National Natural Science Foundation of China (62276230), and the Research Fund for International Scientists of the National Natural Science Foundation of China (72350710798).

\bibliography{custom}

\appendix

\section{Benchmarks and Metrics}
\label{sec:benchmarks}

\paragraph{Datasets and metrics.}
We evaluate the proposed method on both hallucination-focused and general-purpose multimodal benchmarks. The hallucination-focused suite includes CHAIR~\cite{rohrbach2018object}, POPE~\cite{li2023evaluating}, and HallusionBench~\cite{guan2024hallusionbench}, which assess object hallucination, visual grounding, and image-context reasoning from complementary perspectives. To examine whether hallucination mitigation preserves general multimodal capability, we further evaluate on MME~\cite{fu2025mme}, MMBench and MMBench-CN~\cite{liu2024mmbench}, MM-Vet~\cite{yu2023mm}, LLaVA-Bench~\cite{liu2023visual,liu2024llavanext}, and TextVQA~\cite{singh2019towards}. Unless otherwise specified, all results follow the official evaluation protocols of the corresponding benchmarks.

\textbf{CHAIR.}
CHAIR evaluates object hallucination in image captioning by comparing generated object mentions with the ground-truth objects present in the image. We use the MSCOCO val2014 split~\cite{lin2014microsoft}, which provides annotations for 80 object categories, and randomly sample 500 images for evaluation. Each LVLM is prompted with ``Please describe this image in detail.'' CHAIR reports hallucination at both the object-instance and sentence levels:
\begin{equation}
\label{eq:chair}
\begin{aligned}
    \mathrm{CHAIR}_{i} &= |\mathcal{O}_{h}| / |\mathcal{O}_{m}|, \\
    \mathrm{CHAIR}_{s} &= |\mathcal{S}_{h}| / |\mathcal{S}|.
\end{aligned}
\end{equation}
Here, $\mathcal{O}_{h}$ denotes hallucinated object mentions, $\mathcal{O}_{m}$ denotes all mentioned objects, $\mathcal{S}_{h}$ denotes generated sentences containing hallucinated objects, and $\mathcal{S}$ denotes all generated sentences. Lower CHAIR scores indicate fewer hallucinations.

\textbf{POPE.}
POPE is a binary VQA-style benchmark designed to evaluate object hallucination in LVLMs. Each question asks whether a specific object appears in the image, typically in the form ``Is [object] in this image?'' To enforce a clear binary response format, we append a yes-or-no instruction to each question. Following the standard protocol, queried objects are sampled under random, popular, and adversarial settings, and results are reported separately for each setting.

\textbf{HallusionBench.}
HallusionBench evaluates hallucination-related failures in LVLMs under challenging image-context reasoning scenarios. It focuses on two major error sources: language hallucination, where the model relies on linguistic priors that conflict with visual evidence, and visual illusion, where misleading visual cues lead to incorrect predictions. The benchmark contains 346 images and 1,129 expert-written questions, enabling fine-grained evaluation of whether model responses are grounded in both the image content and the textual prompt.

\textbf{MME.}
MME provides a comprehensive evaluation of multimodal perception and cognition. The perception split covers visual understanding tasks such as object existence, counting, spatial localization, color recognition, scene understanding, landmark recognition, artwork identification, and OCR. The cognition split evaluates higher-level abilities, including commonsense reasoning, numerical calculation, translation, and code-related reasoning. Each sample is formulated as a binary question, and the model is required to answer yes or no.

\textbf{MMBench and MMBench-CN.}
MMBench evaluates general visual understanding and reasoning under a unified multiple-choice VQA protocol. It covers a broad range of abilities, including object and attribute recognition, spatial relation understanding, scene comprehension, commonsense reasoning, and text-aware visual reasoning. MMBench-CN follows the same evaluation framework in Chinese, providing a complementary assessment of cross-lingual visual grounding and instruction following. We report accuracy under the official evaluation setting.

\textbf{MM-Vet.}
MM-Vet assesses open-ended multimodal reasoning across diverse capability dimensions, including spatial understanding, attribute and relation reasoning, compositional reasoning, multi-step inference, and text-rich visual reasoning. Since answers are free-form, model outputs are evaluated by an automatic judging pipeline that compares generated responses with reference answers and assigns graded scores according to correctness and instruction following.

\textbf{LLaVA-Bench.}
LLaVA-Bench evaluates visual instruction-following ability in open-ended scenarios, including multi-turn dialogue, detailed image description, and visual reasoning. It contains 60 image-question pairs sampled from natural images. Model responses are compared with reference answers and scored according to the benchmark protocol, providing an additional measure of whether the proposed method preserves general instruction-following ability.

\textbf{TextVQA.}
TextVQA evaluates the ability of LVLMs to read and reason over scene text in real-world images. Questions often require extracting text from signs, menus, product labels, posters, and other text-rich regions, and then grounding the extracted text in the visual context. We report the standard VQA-style accuracy based on answer matching.

\section{Length-Normalized CHAIR Analysis}
\label{sec:app_chair_norm}

The standard CHAIR results in Table~\ref{tab:chair} show that DAC substantially reduces object hallucination. Since DAC also produces slightly shorter captions, we further examine whether this reduction is merely caused by decreased output length. To this end, we report length-normalized hallucinated-object rates on CHAIR.

For the $n$-th generated caption, let $h_n$ denote the number of hallucinated object mentions and $T_n$ denote the number of generated tokens. We define the corpus-level micro rate as
\begin{equation}
\mathrm{HObj}@100_{\mu}
=
100 \cdot
\frac{\sum_{n=1}^{N} h_n}
     {\sum_{n=1}^{N} T_n}.
\label{eq:hobj_micro}
\end{equation}
The sample-level macro rate is defined as
\begin{equation}
\mathrm{HObj}@100_{M}
=
\frac{100}{N}
\sum_{n=1}^{N}
\frac{h_n}{T_n}.
\label{eq:hobj_macro}
\end{equation}
Here, $\mathrm{HObj}@100_{\mu}$ measures hallucinated object mentions per 100 generated tokens over the whole evaluation set, while $\mathrm{HObj}@100_M$ first computes this rate for each sample and then averages across samples.

As shown in Table~\ref{tab:chair_norm}, DAC reduces both normalized hallucination rates. $\mathrm{HObj}@100_{\mu}$ decreases from 1.1206 to 0.5797, and $\mathrm{HObj}@100_M$ decreases from 1.0314 to 0.6569. These results indicate that the CHAIR improvement is not simply an artifact of shorter outputs. Instead, DAC more effectively suppresses unsupported object mentions after controlling for generation length.

\begin{table}[t]
\centering
\caption{Length-normalized CHAIR analysis on LLaVA-1.5. HObj@100$_{\mu}$ measures hallucinated object mentions per 100 generated tokens at the corpus level, while HObj@100$_M$ averages this rate over samples.}
\label{tab:chair_norm}
\small
\renewcommand{\arraystretch}{1.08}
\setlength{\tabcolsep}{0pt}
\begin{tabular*}{\columnwidth}{@{\extracolsep{\fill}}lcc@{}}
\toprule
\textbf{Method} 
& \textbf{HObj@100$_{\mu}$}$\downarrow$ 
& \textbf{HObj@100$_M$}$\downarrow$ \\
\midrule
Regular & 1.1206 & 1.0314 \\
\textbf{DAC} & \textbf{0.5797} & \textbf{0.6569} \\
\bottomrule
\end{tabular*}
\end{table}

\section{Backbones, Baselines, and Evaluation Protocol}
\label{sec:backbones_baselines}

\paragraph{Backbones.}
We evaluate DAC on both established and recent open-source LVLM backbones. The main experiments use LLaVA-1.5~\cite{liu2023visual} and Qwen-VL-Chat as representative backbones, following common practice in hallucination-mitigation evaluation. To further assess generality across newer architectures and model scales, we evaluate DAC on Qwen2.5-VL-3B, Qwen2.5-VL-7B, Qwen3-VL-4B~\cite{wang2024qwen2,bai2025qwen3}, and LLaVA-NeXT-7B~\cite{liu2024llavanext}. This set covers different model families, generations, and parameter scales, allowing us to examine whether DAC is tied to a specific backbone or remains effective for stronger recent LVLMs.

\paragraph{Baselines.}
We compare DAC with regular greedy decoding and representative training-free hallucination mitigation methods. VCD~\cite{leng2024mitigating} and ICD~\cite{wang2024mitigating} are logits- or decoding-level contrastive methods that calibrate token distributions during generation. OPERA~\cite{huang2024opera} is an attention-level method that mitigates hallucination through over-trust penalty and retrospection allocation. MemVR~\cite{zou2024look} is a hidden-state-based method that performs visual retracing in memory space. These baselines cover the major classes of training-free interventions, including logits-level, attention-level, and representation-level mitigation. For completeness, we also include directly comparable reported results from recent methods such as AGLA~\cite{an2025mitigating}, DeGF~\cite{zhang2025self}, VISTA~\cite{pmlr-v267-li25ca}, and AIR~\cite{zhu2026look} when available. Entries that are unsupported by a method or incompatible with our implementation environment are marked as unavailable.

For all methods reimplemented in our experiments, we follow the official implementations and recommended hyperparameters whenever possible. Regular decoding and DAC use greedy decoding unless otherwise specified. Baseline-specific decoding configurations are provided in the reproducibility section. OPERA is omitted on Qwen-based backbones because its implementation depends on older versions of PyTorch and Transformers that are incompatible with our Qwen evaluation pipeline.

\paragraph{Evaluation protocol.}
We evaluate all models on nine multimodal benchmarks: POPE~\cite{li2023evaluating}, CHAIR~\cite{rohrbach2018object}, HallusionBench~\cite{guan2024hallusionbench}, MME~\cite{fu2025mme}, MM-Vet~\cite{yu2023mm}, MMBench and MMBench-CN~\cite{liu2024mmbench}, TextVQA~\cite{singh2019towards}, and LLaVA-Bench~\cite{liu2023visual,liu2024llavanext}. These benchmarks cover hallucination-focused evaluation as well as general multimodal perception, reasoning, OCR, cross-lingual understanding, and open-ended instruction following. For each benchmark, we use the questions from the official annotation files and only adapt the surface format to the chat template required by each LVLM, such as inserting the corresponding system roles, user roles, and image tokens. No benchmark-specific prompt engineering is introduced for DAC.

For POPE, we follow the official COCO-based protocol and report results under the Random, Popular, and Adversarial splits. For CHAIR, we sample images from the MSCOCO Val2014 split~\cite{lin2014microsoft} and query each model with the fixed prompt, ``Please describe this image in detail,'' so that object hallucinations can be measured from free-form captions. For MMBench and MMBench-CN, we follow the official multiple-choice evaluation setting and report accuracy on the corresponding development splits. MME is evaluated under its standard yes-or-no protocol, with results reported for both perception and cognition. TextVQA is evaluated using the standard VQA-style answer-matching accuracy.

For open-ended benchmarks, we follow the official judging pipelines. MM-Vet scores are obtained through the official online evaluator. LLaVA-Bench and HallusionBench are evaluated using their GPT-based judging protocols, and the same evaluator setting is applied to all methods evaluated by us within each benchmark. This ensures that performance differences primarily reflect decoding behavior rather than differences in evaluation prompts or judging procedures.

\section{Reproducibility}
\label{sec:reproducibility}

\paragraph{Implementation details.}
Unless otherwise specified by a baseline method, we use greedy decoding for regular decoding and DAC. For all benchmarks, questions from the official annotation files are used as prompts, and only the surface format is adapted to the chat template required by each LVLM, such as inserting the corresponding system roles, user roles, and image tokens. No benchmark-specific prompt engineering is introduced for DAC. All benchmark evaluations follow the official protocols unless otherwise stated.

For POPE, we evaluate on the COCO-based split and report results under the Random, Popular, and Adversarial settings. For CHAIR, we use images sampled from the MSCOCO Val2014 split and query each model with the fixed prompt, ``Please describe this image in detail.'' The same image subsets are used across all compared methods. We sample three image sets with different random seeds and report the averaged results. For MM-Vet, we use the official online evaluator. For LLaVA-Bench (in-the-wild) and HallusionBench, we follow the corresponding GPT-based judging pipelines and apply the same evaluator setting to all methods evaluated by us.

\paragraph{Reimplemented baselines.}
For all reimplemented baselines, we follow the official implementations and recommended hyperparameters whenever available. VCD~\cite{leng2024mitigating} is implemented with visual contrastive decoding over the original and distorted visual inputs. Following the official implementation, we set \texttt{do\_sample=True}, \texttt{temperature=1}, \texttt{cd\_alpha=1}, \texttt{cd\_beta=0.1}, and \texttt{noise\_step=500}. 

For ICD~\cite{wang2024mitigating}, we follow the official contrastive decoding configuration. The contrastive penalty is set to $\lambda=1$, and the adaptive plausibility constraint is set to $\alpha_{\mathrm{ICD}}=0.1$. We use sampling from the modified softmax distribution with \texttt{temperature=1}, \texttt{top\_p=1}, repetition penalty $1$, and beam size $1$.

For OPERA~\cite{huang2024opera}, we use the official beam-search setting with \texttt{do\_sample=False}, beam size $5$, \texttt{scale\_factor=50}, \texttt{threshold=15}, \texttt{num\_attn\_candidates=5}, and \texttt{penalty\_weights=1}. OPERA is omitted on Qwen-based backbones because its implementation depends on older versions of PyTorch and Transformers that are incompatible with our Qwen evaluation pipeline.

For MemVR~\cite{zou2024look}, we follow the official visual-retracing configuration with \texttt{retracing\_ratio=0.12}, \texttt{entropy\_threshold=0.75}, \texttt{starting\_layer=5}, and \texttt{ending\_layer=16}. We use greedy decoding with \texttt{do\_sample=False}, \texttt{temperature=0}, and beam size $1$ unless the official script specifies otherwise.

\paragraph{Reported baselines.}
In addition to the methods reimplemented above, we include directly comparable reported results from recent training-free mitigation methods, including AGLA~\cite{an2025mitigating}, DeGF~\cite{zhang2025self}, VISTA~\cite{pmlr-v267-li25ca}, and AIR~\cite{zhu2026look}, when results are available under the same backbone and benchmark protocol. These entries are taken from the corresponding papers or official result tables without additional re-tuning. Missing entries indicate that the method does not support the corresponding setting or that directly comparable results are not publicly available. These reported baselines are not included in the inference-efficiency comparison.

\paragraph{DAC configuration.}
For DAC, we use greedy decoding with \texttt{do\_sample=False}, \texttt{temperature=0}, and beam size $1$, following the regular decoding setting. The DAC hyperparameters are fixed for all datasets: $\gamma=0.5$ and $\alpha=0.2$ for Layer-wise Semantic Compensation, and $\tau=0.3$ and $\beta=0.3$ for Sequential Semantic Correction. We do not perform dataset-specific hyperparameter tuning.

\paragraph{Computational resources.}
All experiments are conducted on NVIDIA RTX 4090 GPUs. For efficiency evaluation, all methods are tested under the same hardware environment using LLaVA-1.5 on MM-Vet, and we report per-token latency, decoding throughput, total time for generating 80 tokens, and peak GPU memory usage.

\section{The Algorithm of DAC}
\label{sec:app_algorithm}

In this section, we present the full algorithmic details of DAC.

\begin{algorithm}[t!]
\caption{Dynamic Alignment Compensation (DAC)}
\label{alg:dac}
\footnotesize
\begin{algorithmic}[1]
\REQUIRE LVLM $\mathcal{M}_{\Theta}$ with decoder blocks $\{\mathcal{F}_l\}_{l=1}^{L}$
\REQUIRE Image $I$, query $Q$
\REQUIRE Thresholds $\gamma,\tau$; coefficients $\alpha,\beta$
\ENSURE Generated sequence $y_{1:T}$

\STATE Obtain visual tokens $X_{\mathrm{vis}}$ from $I$
\STATE Obtain textual tokens $X_{\mathrm{txt}}$ from $Q$
\STATE $X_0 \gets [X_{\mathrm{vis}},X_{\mathrm{txt}}]$
\STATE Initialize temporal caches $c_l \gets \emptyset$, $l=1,\ldots,L$
\STATE Let $s(\cdot)$ denote feature-wise softmax

\FOR{$t=1$ \textbf{to} $T$}
    \STATE $X_t \gets [X_0,y_{<t}]$
    \STATE $H_{0,t} \gets \operatorname{Embed}(X_t)$

    \FOR{$l=1$ \textbf{to} $L$}
        \STATE $H_{l,t} \gets \mathcal{F}_l(H_{l-1,t})$

        \IF{$l>1$}
            \FOR{each token $i$ in $X_t$}
                \STATE $p_{l-1,t}^{(i)} \gets s\!\left(H_{l-1,t}^{(i)}\right)$
                \STATE $p_{l,t}^{(i)} \gets s\!\left(H_{l,t}^{(i)}\right)$
                \STATE $D_{l,t}^{(i)} \gets 
                \mathrm{JS}\!\left(
                p_{l-1,t}^{(i)} \,\|\, p_{l,t}^{(i)}
                \right)$
                \IF{$D_{l,t}^{(i)} > \gamma$}
                    \STATE $H_{l,t}^{(i)} \gets 
                    H_{l,t}^{(i)} + \alpha H_{l-1,t}^{(i)}$
                \ENDIF
            \ENDFOR
        \ENDIF

        \STATE $h_{l,t} \gets H_{l,t}^{(\mathrm{cur})}$
        \IF{$c_l \neq \emptyset$}
            \STATE $q_{l,t} \gets s(h_{l,t})$
            \STATE $q_{l,t-1} \gets s(c_l)$
            \STATE $\Delta_{l,t} \gets 
            \mathrm{JS}\!\left(q_{l,t} \,\|\, q_{l,t-1}\right)$
            \IF{$\Delta_{l,t} > \tau$}
                \STATE $h_{l,t} \gets 
                h_{l,t} + \beta(c_l-h_{l,t})$
                \STATE $H_{l,t}^{(\mathrm{cur})} \gets h_{l,t}$
            \ENDIF
        \ENDIF

        \STATE $c_l \gets h_{l,t}$
    \ENDFOR

    \STATE $y_t \gets \operatorname{Decode}\!\left(H_{L,t}^{(\mathrm{cur})}\right)$
\ENDFOR
\end{algorithmic}
\end{algorithm}

\section{Examples of Capability Integrations}
\begin{figure*}[h]
    \centering
    \includegraphics[width=1.0\linewidth]{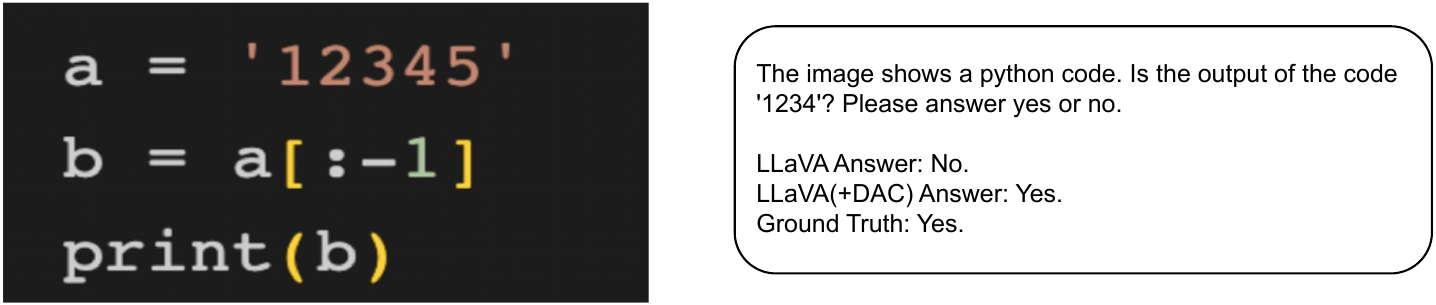}
    \caption{A case study comparing the levels of hallucination among various baselines}
    \label{fig:placeholder}
\end{figure*}

\begin{figure*}[h]
    \centering
    \includegraphics[width=1.0\linewidth]{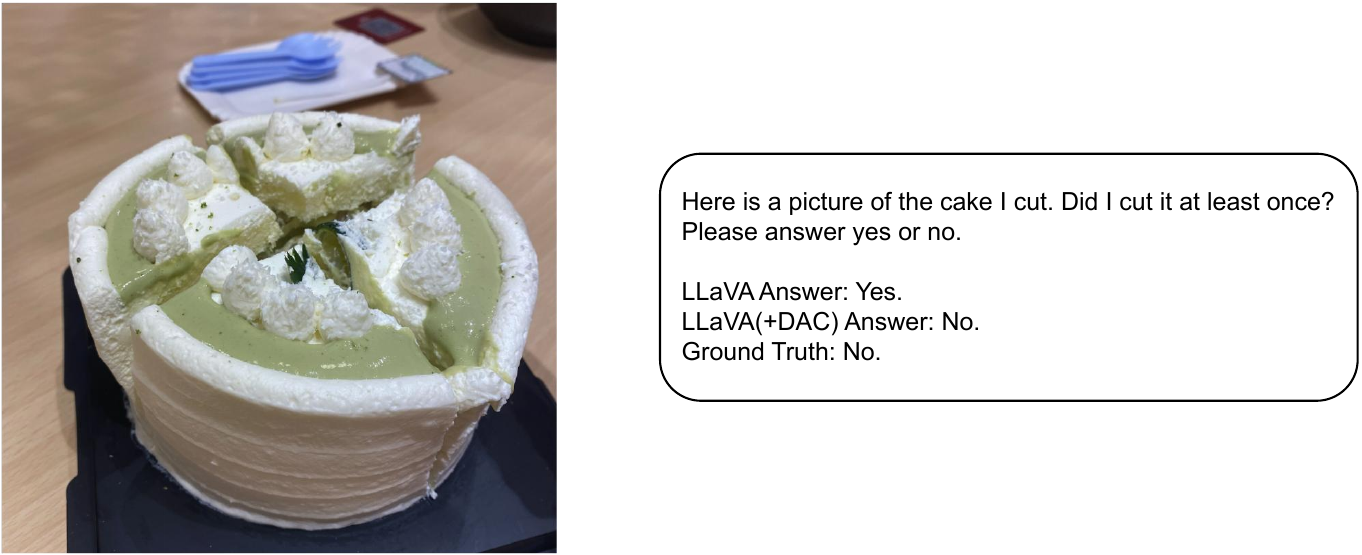}
    \caption{A case study comparing the levels of hallucination among various baselines}
\end{figure*}

\begin{figure*}[h]
    \centering
    \includegraphics[width=1.0\linewidth]{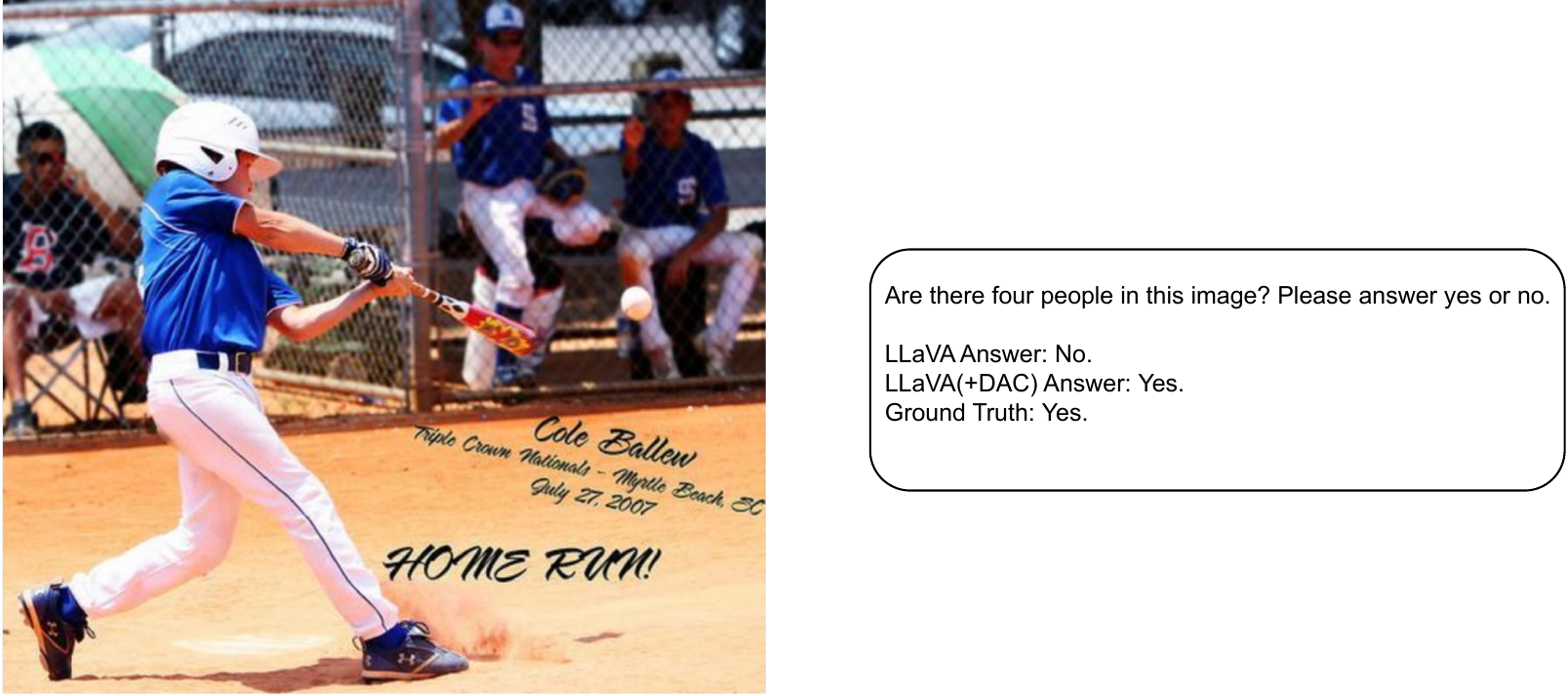}
    \caption{A case study comparing the levels of hallucination among various baselines}
\end{figure*}

\begin{figure*}[h]
    \centering
    \includegraphics[width=1.0\linewidth]{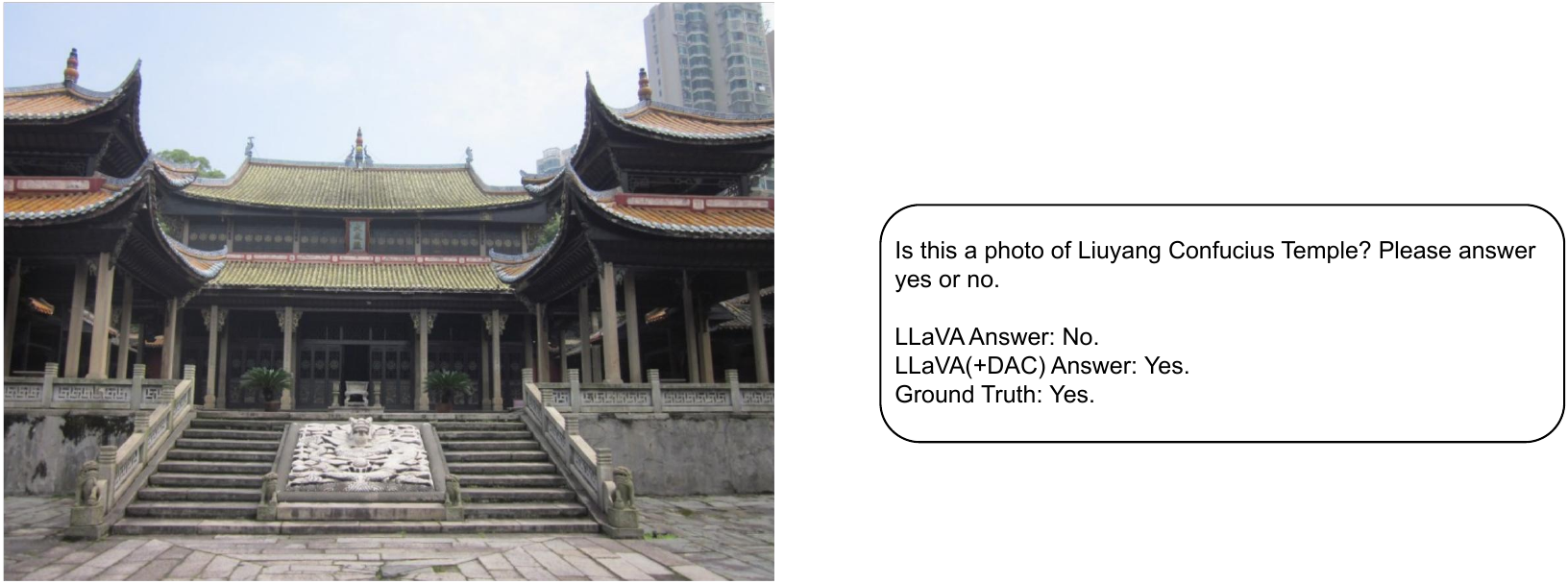}
    \caption{A case study comparing the levels of hallucination among various baselines}
\end{figure*}

\begin{figure*}[h]
    \centering
    \includegraphics[width=1.0\linewidth]{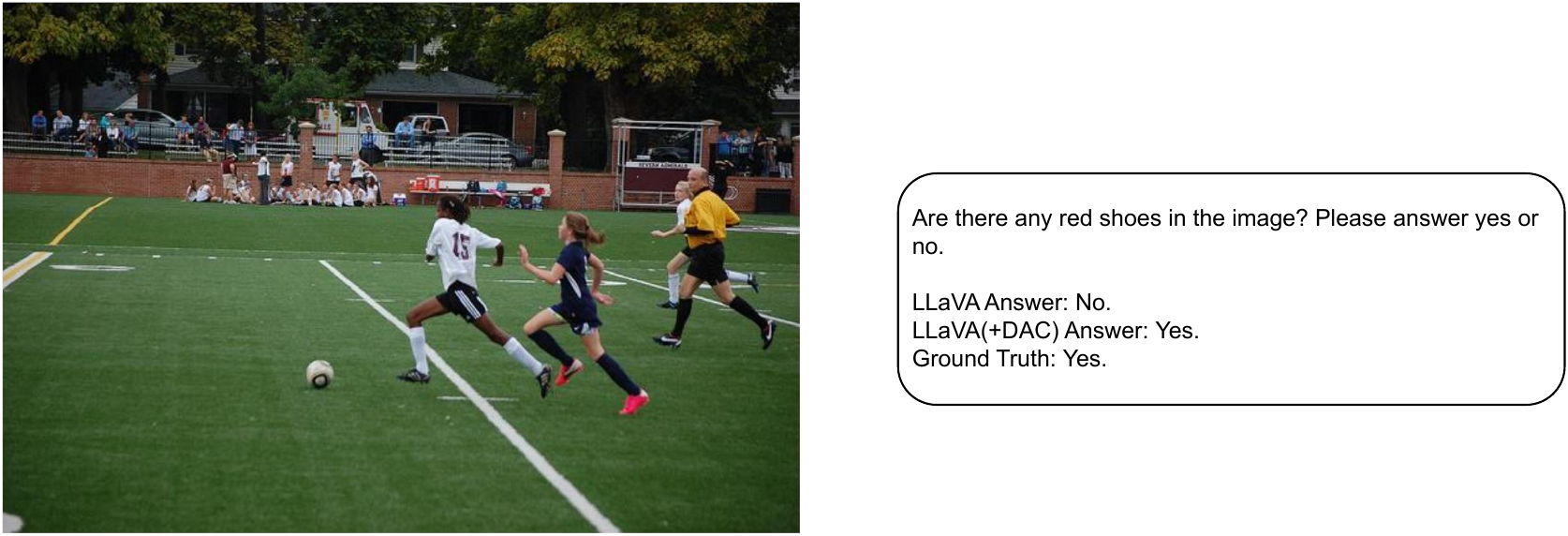}
    \caption{A case study comparing the levels of hallucination among various baselines}
\end{figure*}

\begin{figure*}[h]
    \centering
    \includegraphics[width=1.0\linewidth]{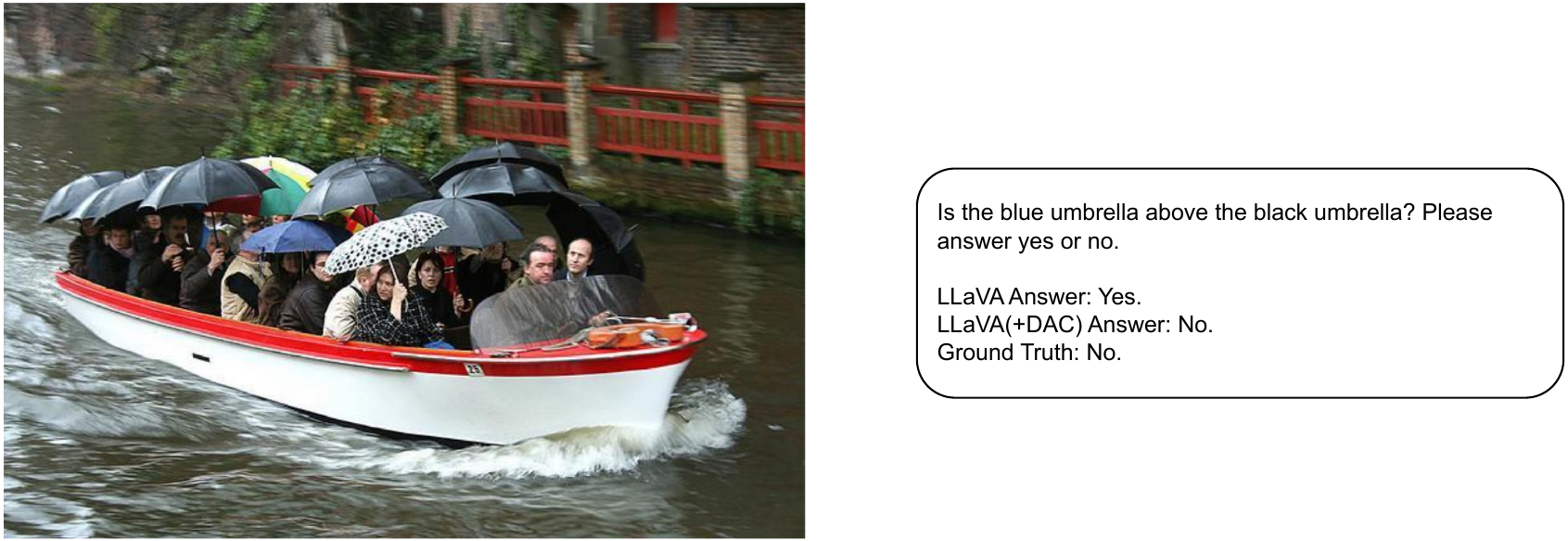}
    \caption{A case study comparing the levels of hallucination among various baselines}
\end{figure*}

\begin{figure*}[h]
    \centering
    \includegraphics[width=1.0\linewidth]{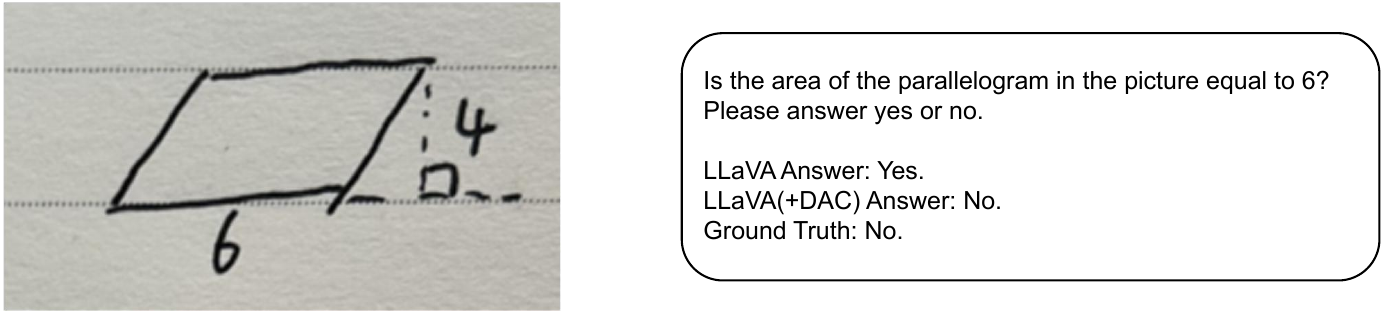}
    \caption{A case study comparing the levels of hallucination among various baselines}
\end{figure*}

\begin{figure*}[h]
    \centering
    \includegraphics[width=1.0\linewidth]{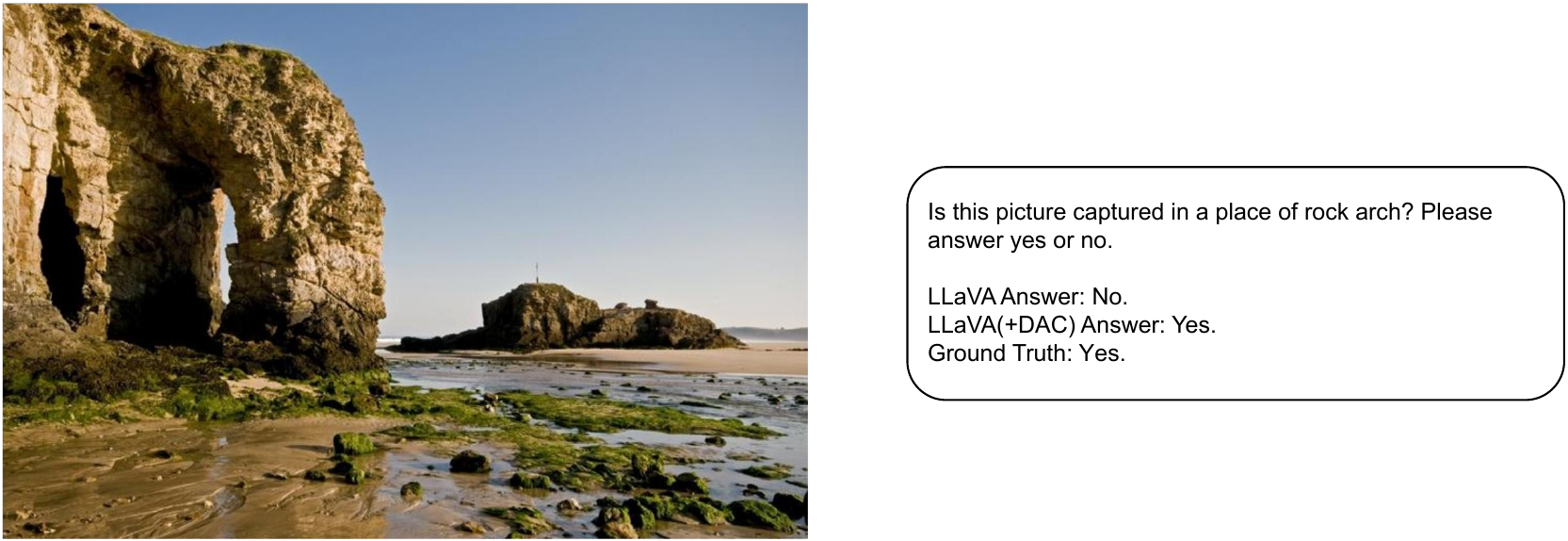}
    \caption{A case study comparing the levels of hallucination among various baselines}
\end{figure*}

\begin{figure*}[h]
    \centering
    \includegraphics[width=1.0\linewidth]{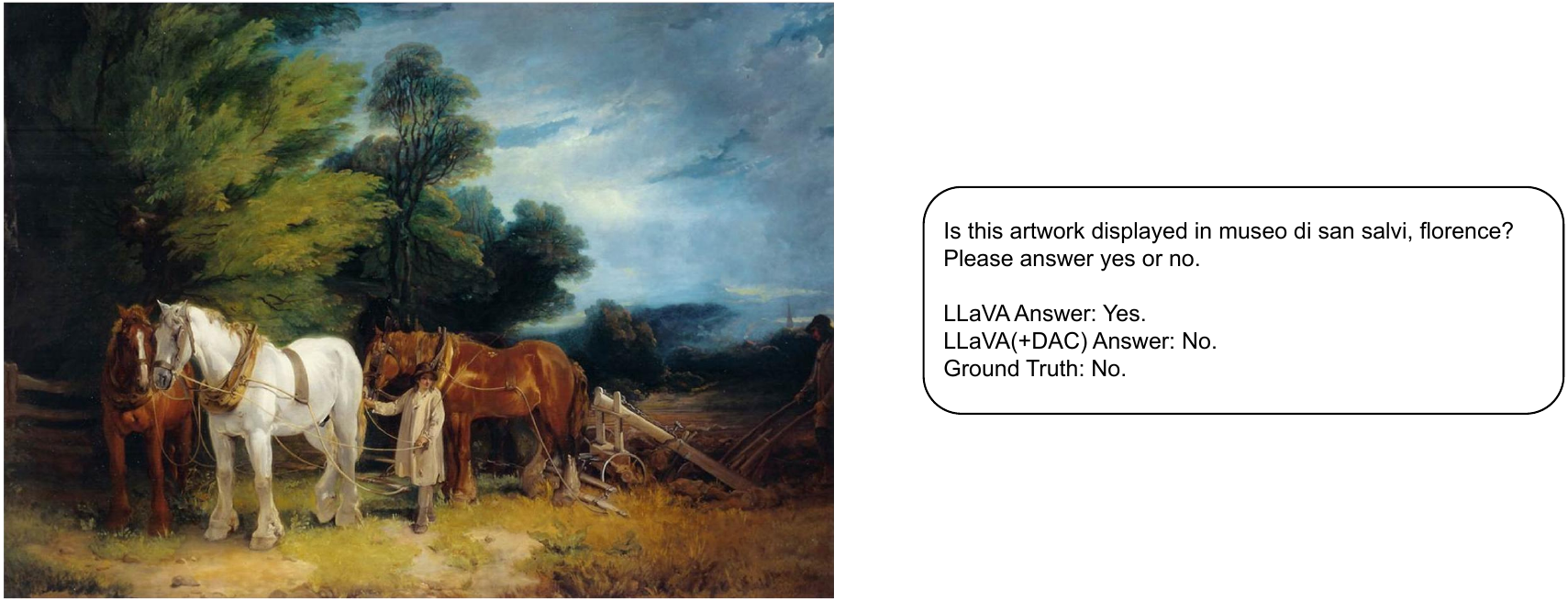}
    \caption{A case study comparing the levels of hallucination among various baselines}
\end{figure*}

\begin{figure*}[h]
    \centering
    \includegraphics[width=1.0\linewidth]{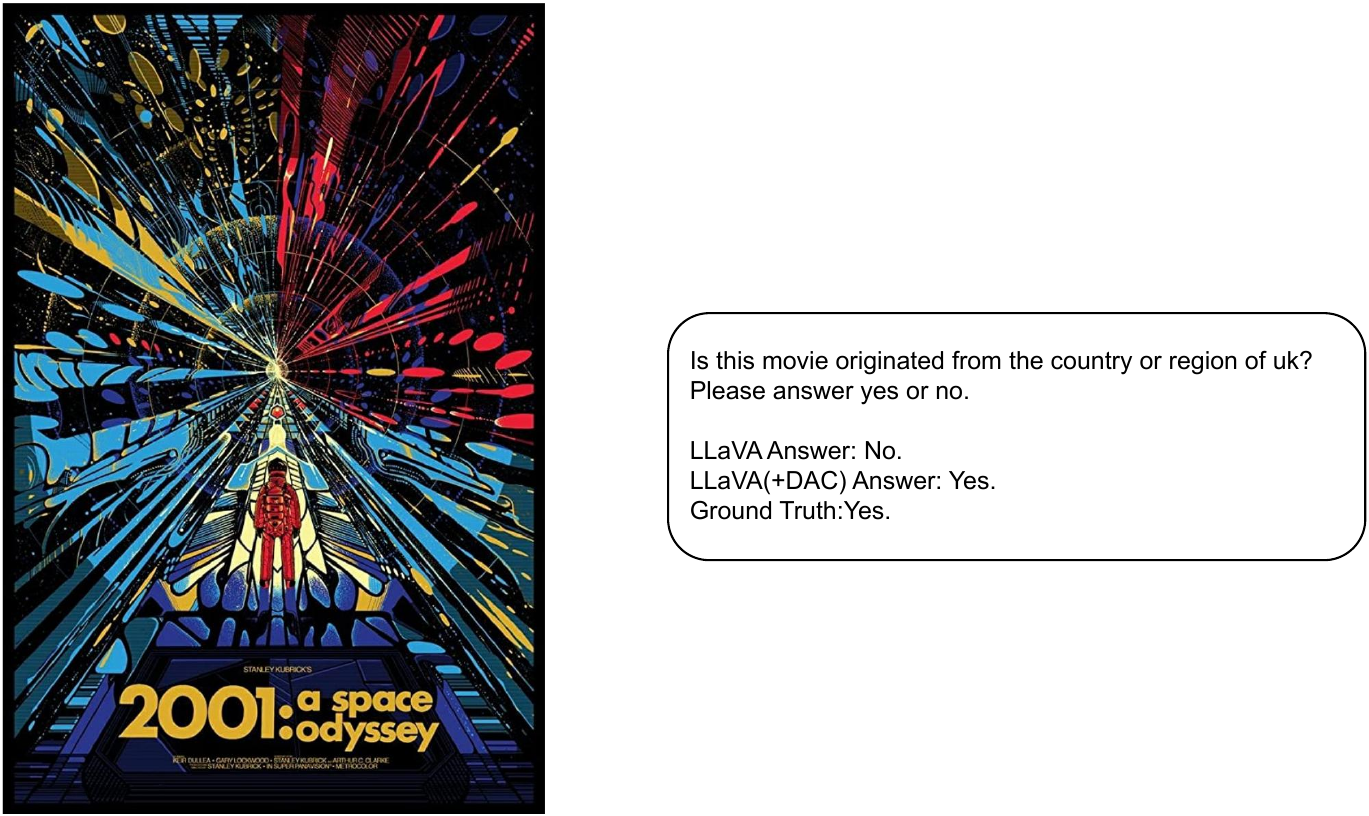}
    \caption{A case study comparing the levels of hallucination among various baselines}
\end{figure*}

\begin{figure*}[h]
    \centering
    \includegraphics[width=1.0\linewidth]{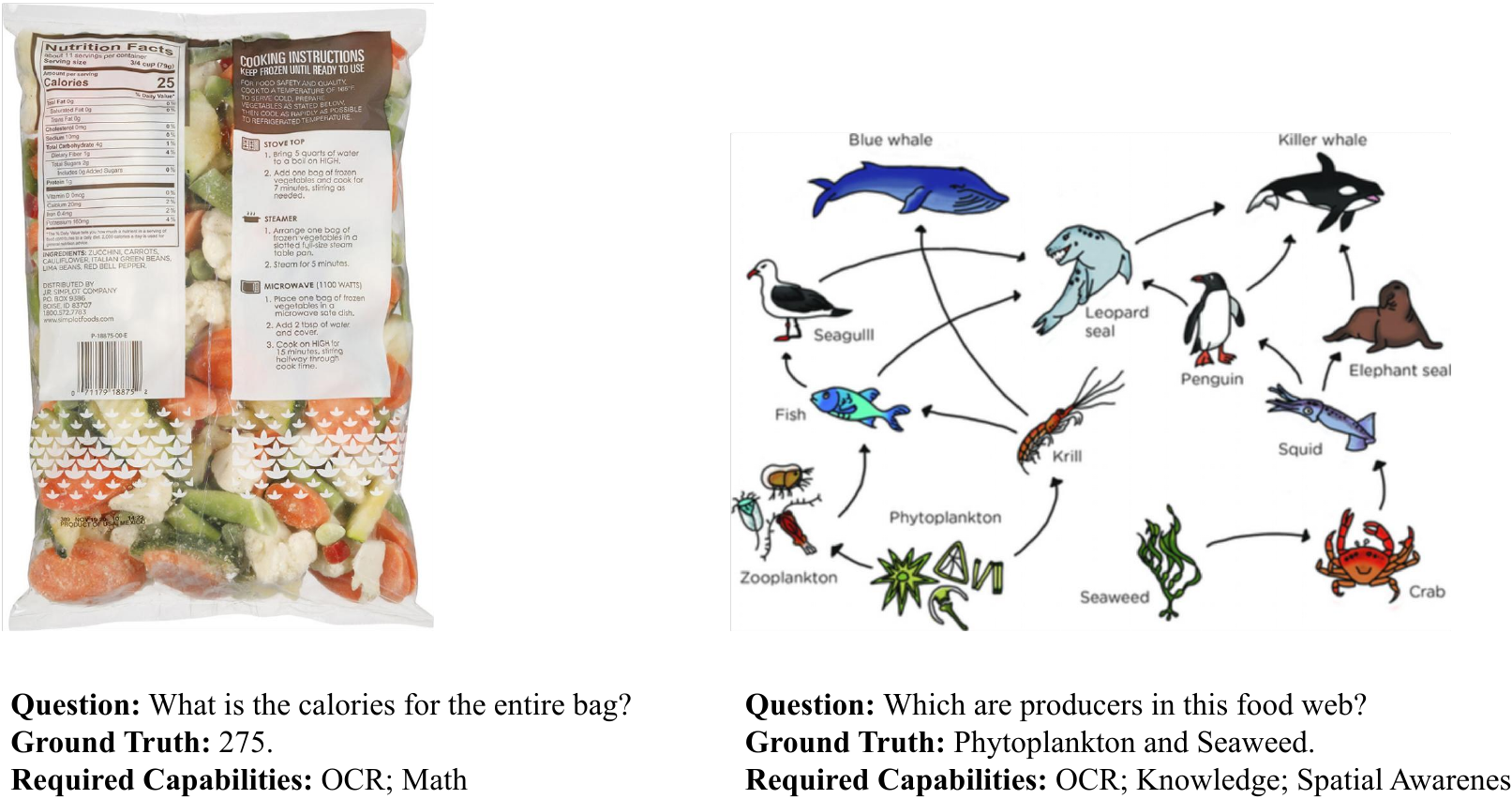}
    \caption{A case study comparing the levels of hallucination among various baselines}
\end{figure*}

\begin{figure*}[h]
    \centering
    \includegraphics[width=1.0\linewidth]{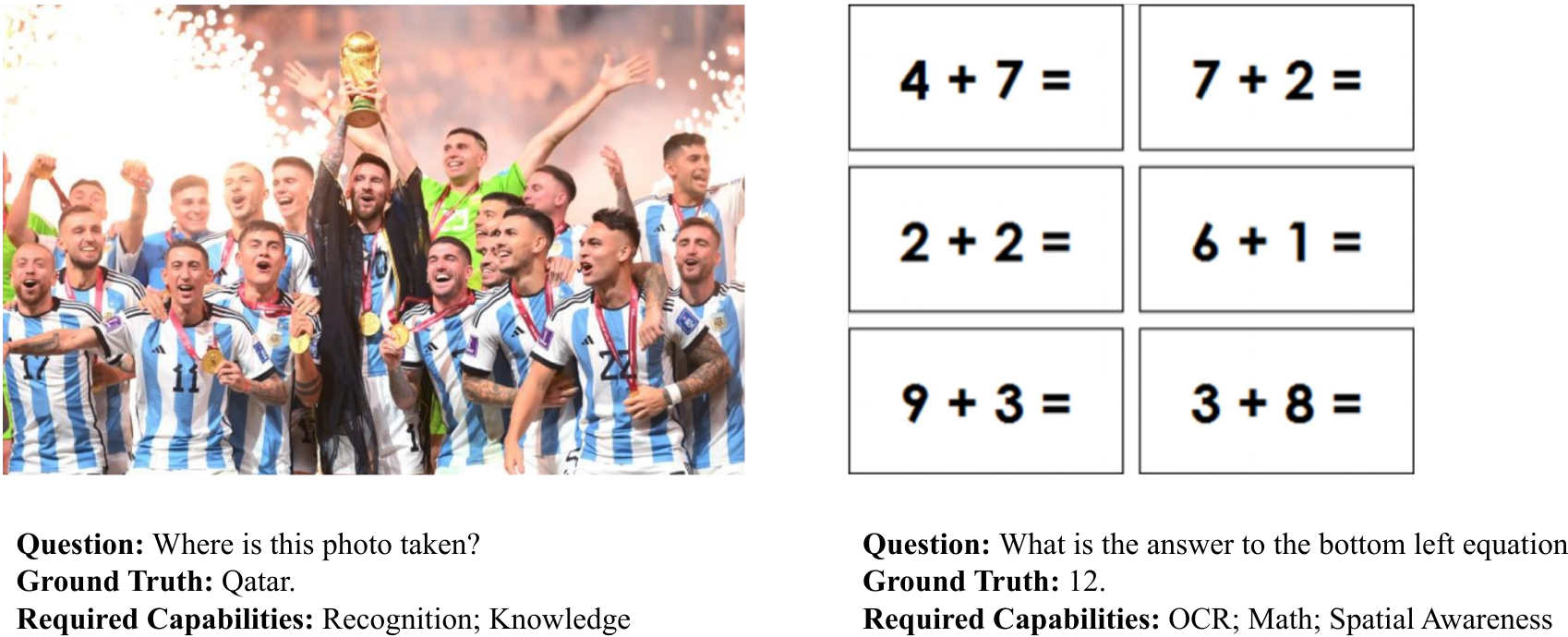}
    \caption{A case study comparing the levels of hallucination among various baselines}
\end{figure*}

\begin{figure*}[h]
    \centering
    \includegraphics[width=1.0\linewidth]{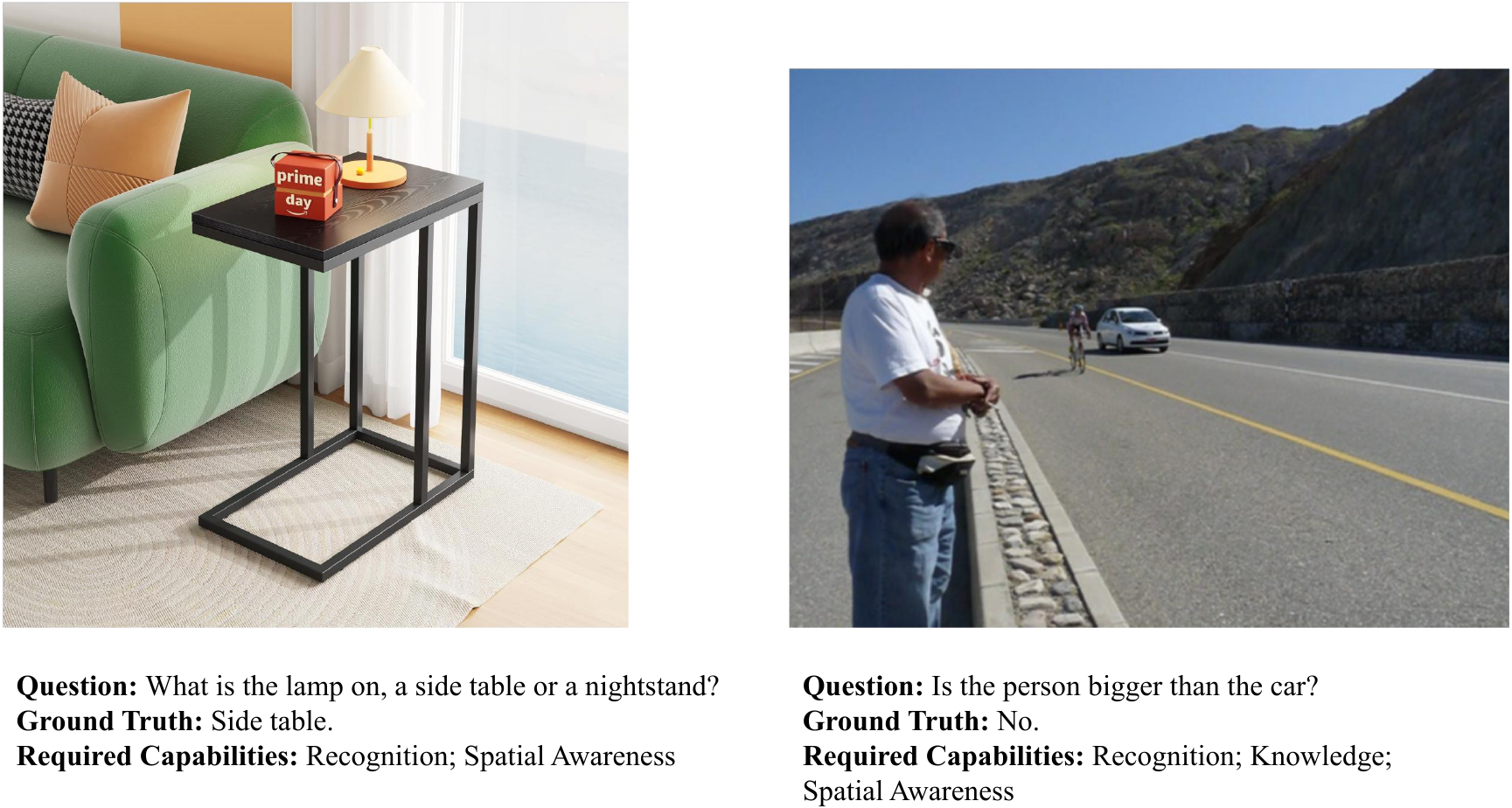}
    \caption{A case study comparing the levels of hallucination among various baselines}
\end{figure*}
\end{document}